\documentclass{article}

\usepackage[preprint]{neurips_2025}

\usepackage{fontawesome}
\usepackage{pifont}
\usepackage[utf8]{inputenc} % allow utf-8 input
\usepackage[T1]{fontenc}    % use 8-bit T1 fonts
\usepackage{hyperref}       % hyperlinks
\usepackage{url}            % simple URL typesetting
\usepackage{booktabs}       % professional-quality tables
\usepackage{amsfonts}       % blackboard math symbols
\usepackage{nicefrac}       % compact symbols for 1/2, etc.
\usepackage{microtype}      % microtypography
\usepackage{xcolor}         % colors
\usepackage[most]{tcolorbox}
\usepackage{amsthm}  % 确保引入该宏包
\theoremstyle{plain}

\theoremstyle{definition}
\newtheorem{definition}{Definition}

\theoremstyle{remark}

\usepackage{wrapfig}
\usepackage{graphicx}
\usepackage{amsmath}
\usepackage{amssymb}
\usepackage{bbm}
\usepackage{multirow} 
\usepackage{array}
\usepackage{tabularx}
\usepackage{color, colortbl}
\definecolor{LightCyan}{rgb}{0.88,1,1}
\definecolor{LightGrey}{rgb}{0.95, 0.95, 0.95}

\usepackage{mathtools}
\usepackage{wrapfig}
\usepackage{lipsum}
\usepackage{mwe}
\usepackage{subfig} 
\usepackage{algorithm}
\usepackage{algpseudocode}
\algrenewcommand\algorithmicrequire{\textbf{Input:}}
\algrenewcommand\algorithmicensure{\textbf{Output:}}

\usepackage{enumitem}

\usepackage[utf8]{inputenc} % allow utf-8 input
\usepackage[T1]{fontenc}    % use 8-bit T1 fonts
\usepackage{hyperref}       % hyperlinks
\usepackage{url}            % simple URL typesetting
\usepackage{booktabs}       % professional-quality tables
\usepackage{amsfonts}       % blackboard math symbols
\usepackage{nicefrac}       % compact symbols for 1/2, etc.

\usepackage{hyperref}
\usepackage{natbib}
\definecolor{seal}{HTML}{B711AC}
\definecolor{applegreen}{HTML}{66c2a5}

\definecolor{groupbg}{gray}{0.93}

\title{\raisebox{-1ex}{\includegraphics[height=3ex]{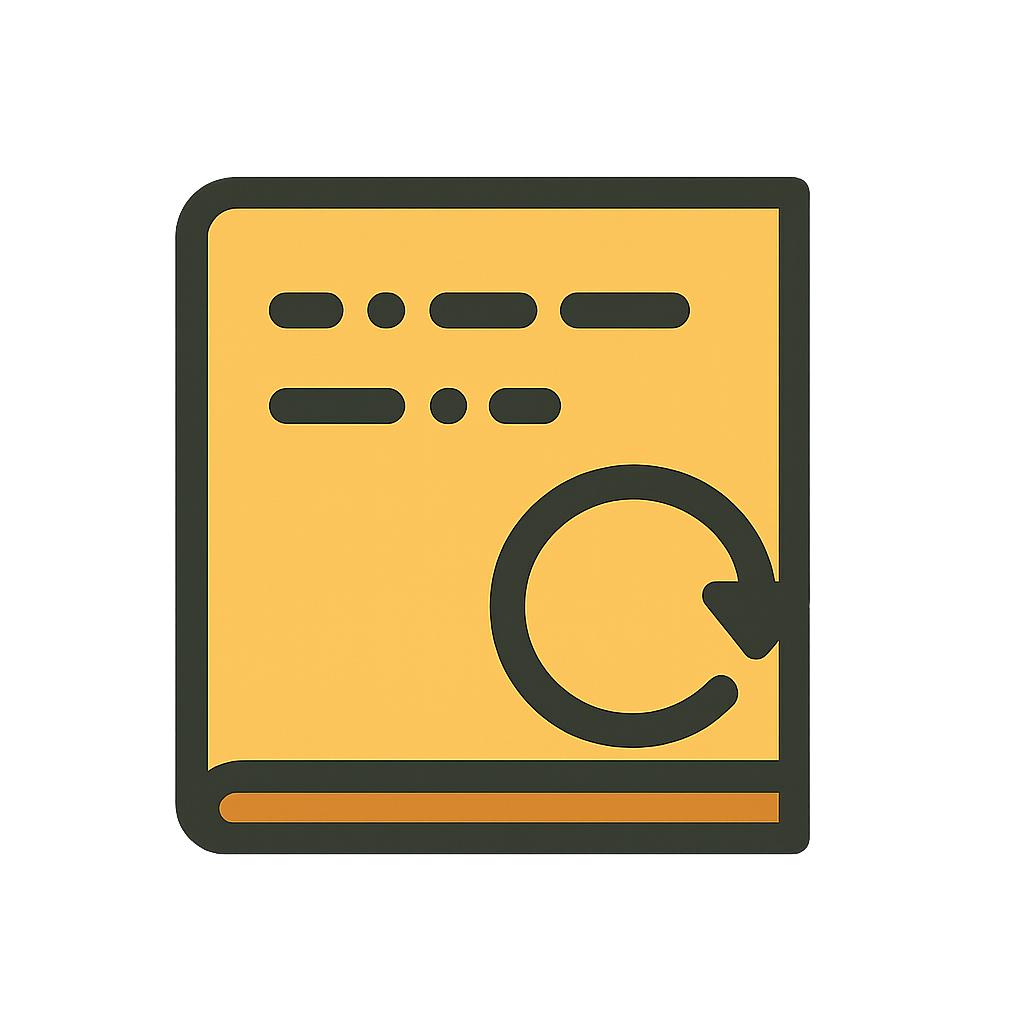}}CodeMerge: Codebook-Guided Model Merging for Robust Test-Time Adaptation in Autonomous Driving}

\author{%
Huitong Yang \quad Zhuoxiao Chen  \quad Fengyi Zhang  \quad \textbf{Zi Huang} \quad \textbf{Yadan Luo} \\
  The University of Queensland\\
  \texttt{\{huitong.yang, zhuoxiao.chen, fengyi.zhang, helen.huang, y.luo\}@uq.edu.au} 
}

\begin{document}

\maketitle

\begin{abstract}
Maintaining robust 3D perception under dynamic and unpredictable test-time conditions remains a critical challenge for autonomous driving systems. Existing test-time adaptation (TTA) methods often fail in high-variance tasks like 3D object detection due to unstable optimization and sharp minima. While recent model merging strategies based on linear mode connectivity (LMC) offer improved stability by interpolating between fine-tuned checkpoints, they are computationally expensive, requiring repeated checkpoint access and multiple forward passes. In this paper, we introduce CodeMerge, a lightweight and scalable model merging framework that bypasses these limitations by operating in a compact latent space. Instead of loading full models, CodeMerge represents each checkpoint with a low-dimensional fingerprint derived from the source model’s penultimate features and constructs a key-value codebook. We compute merging coefficients using ridge leverage scores on these fingerprints, enabling efficient model composition without compromising adaptation quality. Our method achieves strong performance across challenging benchmarks, improving end-to-end 3D detection 14.9\% NDS on nuScenes-C and LiDAR-based detection by over 7.6\% mAP on nuScenes-to-KITTI, while benefiting downstream tasks such as online mapping, motion prediction and planning even without training. Code and pretrained models are released in the supplementary material.
\end{abstract}

\vspace{-2ex}
\section{Introduction}
\begin{figure*}[t]\vspace{-2ex}
    \includegraphics[width=1\linewidth]{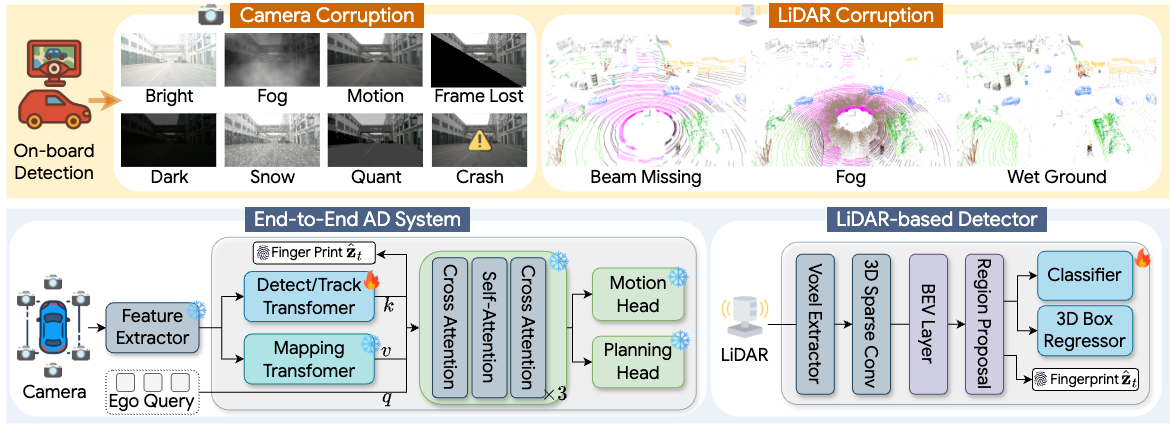}\vspace{-1ex}
    \caption{\textbf{Overview of real-world test-time shifts (top) and 3D perception systems considered in this work (bottom).} We study test-time adaptation (TTA) in two settings: (1) an end-to-end autonomous driving system and (2) a modular LiDAR-based detector, both affected by adverse weather and sensor failures. CodeMerge enables efficient TTA by leveraging compact fingerprints to guide model merging.\vspace{-2ex}}\label{fig:teaser}\vspace{-1ex}
\end{figure*}

\vspace{-1ex}
Real-world autonomous driving scenarios often encounter rapid and unpredictable environmental variations, such as sudden adverse weather conditions \textit{(e.g.,} fog, snow) or sensor malfunctions (\textit{e.g.,} dropped frames, missing beams) arising from LiDAR and camera systems, as illustrated in Fig. \ref{fig:teaser}. These abrupt disruptions momentarily render 3D perception modules partially or fully ``blind'', propagating erroneous decision-making downstream and leading to severe safety hazards in the end-to-end autonomous driving (AD) pipeline. Consequently, a critical yet unresolved question emerges: \textit{Can perception models efficiently and robustly adapt onboard to address such unforeseen distributional shifts?}

Test-time adaptation (TTA) offers a promising direction by enabling models to adapt online during inference. Prior TTA approaches typically handle shifts by aligning BatchNorm statistics \citep{DBLP:conf/iclr/WangSLOD21, DBLP:conf/iclr/Niu00WCZT23}, enforcing consistency through data augmentations \citep{DBLP:conf/cvpr/0013FGD22}, or minimizing sharpness via adversarial perturbations \citep{gong2024sotta, DBLP:conf/iclr/Niu00WCZT23}. Nonetheless, when directly extending them to complex tasks such as 3D detection, these approaches often suffer from brittle optimization dynamics and fall into sharp local minima, which can lead to the loss of previously acquired generalization and the ability to cope with future task distributions \citep{DBLP:conf/mm/ChenWL0H24}.

Recent studies improve \textit{long-term} adaptation stability by leveraging model merging techniques \cite{DBLP:conf/iclr/KimABKLAHSY25} grounded in linear mode connectivity (LMC), which posits that models fine-tuned on different target samples but initialized from \textit{the same} pretrained source model are \textit{``linearly connected''} in weight space. Thus, interpolating between such models has been shown to produce reliable pseudo-labels and alleviate model collapse issues in TTA \citep{DBLP:conf/iclr/Niu00WCZT23}. Techniques such as Mean Teacher and their variants exponentially averages the weights of past models, but often underutilize valuable diversity across past checkpoints. More recently, Model Synergy (MOS) \citep{DBLP:conf/iclr/ChenMB0HL25} extends this idea by maintaining a buffer of Top-$K$ important checkpoints and dynamically merging them using synergy weights, computed via kernel similarities among each model's predictions of the current test sample. While effective, MOS incurs significant overhead from repeatedly checkpoint loading and performing $K$ forward passes for calculating synergy weights, limiting its scalability in high-throughput driving applications.

In this work, we introduce a \underline{code}book-guided model merging (CodeMerge) approach for adapting 3D perception modules against various shifts at test time. The core idea is to represent each fine-tuned checkpoint $\Phi_{\Theta^{(t)}}$ by a compact ``fingerprint'' derived from the source model's penultimate activations. These fingerprints serve as keys in a model codebook, mapping to their corresponding checkpoint weights. Crucially, correlations in this low-dimensional fingerprint space reliably mirror those in the high-dimensional weight space (see Figure \ref{fig:corr}), enabling informed merging decisions without loading full model parameters. CodeMerge employs ridge leverage scores to rank the informativeness of fingerprints, a technique theoretically linked to approximations of the inverse Hessian in the parameter space.  This procedure needs memory that scales only with the fingerprint dimension and adds negligible latency, yet it lifts end-to-end 3D detection NDS by 14.9\%, tracking AMOTA on the nuScenes-C corruption benchmark by 19.3\%, and LiDAR-based detection 3D mAP by 7.6\% on the challenging nuScenes‑to‑KITTI shift. These improvements seamlessly propagate to downstream motion prediction and planning modules without modification or additional training. Source code is available in the supplementary material.

\vspace{-2ex}
\section{Preliminaries}\vspace{-1ex}
We begin by formalizing the problem setting for test-time adaptation (TTA) in 3D object detection and reviewing model merging strategies that exploit linear mode connectivity in such context.

\noindent\textbf{Task Formulation.} Let $\Phi_{\Theta^{(0)}}=\phi_{\Theta^{(0)}}\circ h_{\Theta^{(0)}}$ denote a pretrained 3D object detection model, comprising a feature extractor $\phi_{\Theta^{(0)}}(\cdot):\mathbf{X}\mapsto \mathbf{Z}\in\mathbb{R}^{d}$ maps an input $\mathbf{x}\in\mathbf{X}$ (\textit{e.g.,} a point cloud or multi-view images) to a latent feature map $\mathbf{z}\in\mathbf{Z}$, and the head regresses 3D boxes $h_{\Theta^{(0)}}(\cdot):\mathbf{Z}\mapsto \mathcal{Y}\in\mathbb{R}^{7}$. The goal of TTA is to sequentially adapt the model to a stream of unlabeled target-domain inputs $\mathcal{D}_{\mathrm{tar}} = \{\mathbf{x}_t\}_{t=1}^{T}$, which may exhibit significant distributional shifts or corruptions. The online adaptation must follow in a single forward-pass setting, incrementally evolving the model parameters $\Theta^{(0)} \rightarrow \Theta^{(1)} \rightarrow \ldots \rightarrow \Theta^{(t)}$ to improve detection over time. 

\begin{figure*}[t]\vspace{-2ex}
    \includegraphics[width=1\linewidth]{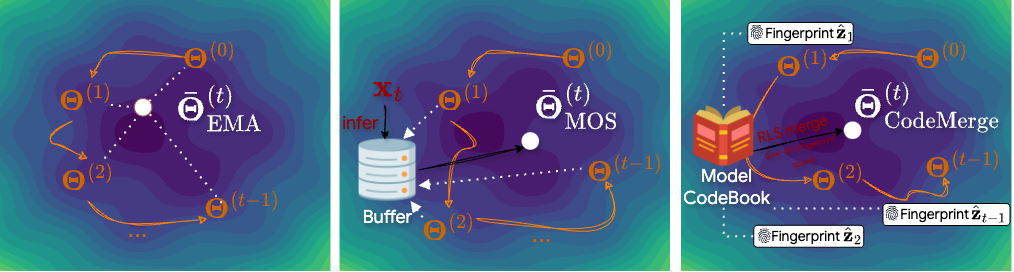}
    \caption{\textbf{Conceptual comparison of model merging strategies for TTA.} Unlike EMA (left), which ignores model behavior, or MOS (middle), which requires multiple inferences to compute merging weights, CodeMerge (right) leverages ridge leverage scores in a compact fingerprint space to efficiently guide model merging.\vspace{-2ex}}\label{fig:model}
\end{figure*}

\noindent\textbf{Linear Mode Connectivity (LMC).} LMC \cite{DBLP:conf/iclr/KimABKLAHSY25, DBLP:conf/icml/FrankleD0C20, DBLP:conf/cvpr/WortsmanIKLKRLH22} refers to the empirical property that two models $\Theta^{(1)}$ and $\Theta^{(2)}$ trained from a shared initialization (or sufficiently close regions in weight space), can be connected by a \textit{``linear path''} without significant loss degradation. Formally, for any $\lambda\in[0, 1]$,
\begin{equation}\label{eq:lmc}
    \mathcal{L}\left((1-\lambda)\Theta^{(1)} + \lambda \Theta^{(2)}\right) \approx (1-\lambda)\mathcal{L}(\Theta^{(1)}) + \lambda \mathcal{L}(\Theta^{(2)}).
\end{equation}
This property facilitates efficient model merging through linear interpolation.

\noindent\textbf{Implication for Model Merging in TTA.}
If LMC holds true between each pair of successive parameters $(\Theta^{(t-1)}, \Theta^{(t)})$ fine-tuned from $\Theta^{(0)}$, then their interpolated model should yield low loss. This underpins methods like \textit{Mean Teacher} shown in Fig. \ref{fig:model}, in which teacher models are recursively updated with an exponential moving average (EMA) with a decay factor $\beta\in(0, 1)$:
\begin{equation}
    \bar{\Theta}^{(t)}_{\mathrm{EMA}} = \beta \bar{\Theta}_{\mathrm{EMA}}^{(t-1)} + (1-\beta) \Theta^{(t)} \Rightarrow\bar{\Theta}^{(t)}_{\mathrm{EMA}} = (1-\beta)\sum_{i=0}^t \beta^{t-i} \Theta^{(i)}.
\end{equation}
 Under LMC, this leads to approximately linear combinations of multi-task losses: \vspace{-1ex}
\begin{equation}
    \mathcal{L}(\bar{\Theta}_{\mathrm{EMA}}^{(t)}) \approx   (1-\beta)\sum_{i=0}^t \beta^{t-i} \mathcal{L}(\Theta^{(i)}).
\end{equation}
This shows that averaging can reduce variance from balancing multi-task losses. However, EMA's coefficients are solely based on time steps rather than model behavior, making it potentially suboptimal.

In contrast, MOS~\cite{DBLP:conf/iclr/ChenMB0HL25} (middle in Fig. \ref{fig:model}) adaptively merges model parameters by solving a kernel-weighted least squares problem over a buffer of $K$ candidate checkpoints $\{\Theta^{(i)}\}_{i=1}^K$. Given a test batch $\mathbf{x}_t$, the merged model is computed as: \vspace{-1ex}\begin{equation}
\bar{\Theta}^{(t)} = \sum_{i=1}^K \tilde{w}_i^{(t)} \Theta^{(i)}, ~\text{where}~ \tilde{w}_i^{(t)} = \frac{\sum_j [\mathbf{K}^{(t)}]^{-1}_{ij}}{\sum_{i', j'} [\mathbf{K}^{(t)}]^{-1}_{i'j'}},
\end{equation}\vspace{-1ex}
\begin{equation}
\mathbf{K}^{(t)}_{ij} = \mathrm{Sim}\left( \Phi_{\Theta^{(i)}}(\mathbf{x}_t), \Phi_{\Theta^{(j)}}(\mathbf{x}_t) \right) \cdot \mathrm{Sim}\left( \phi_{\Theta^{(i)}}(\mathbf{x}_t), \phi_{\Theta^{(j)}}(\mathbf{x}_t) \right),
\end{equation}
where kernel matrix $\mathbf{K}^{(t)} \in \mathbb{R}^{K \times K}$ captures pairwise similarity between model outputs under the current batch. To evaluate $\tilde{w}_i^{(t)}$, MOS requires $K$ forward passes over $\mathbf{x}_t$, making it more computationally intensive and thus hard to scale up the horizon $K$ in TTA.

\vspace{-1ex}
\section{Our Approach}\vspace{-1ex}
We introduce CodeMerge, a codebook-guided model merging scheme for efficient TTA in 3D object detection \textit{without} triggering repeated inference across past models. To achieve this, we construct a model codebook (Sec. \ref{sec:codebook}), where each checkpoint is represented by a compact fingerprint derived from intermediate features of a fixed source model. During inference, we compute curvature-aware ridge leverage scores (Sec. \ref{sec:fingerprints})  in the fingerprint space. Finally, we perform a sign-consistent weighted merge of top-scoring candidate models (Sec. \ref{sec:merge}), promoting both stability and diversity. \vspace{-1ex}

\subsection{Model CodeBook} \label{sec:codebook}\vspace{-1ex}
At each step $t$, we maintain a model codebook for \textit{all} past checkpoints along the adaptation trajectory, denoted as:
\begin{equation}
\mathcal{C}^{(t)}=\{\hat{\mathbf{z}}_{i}:\Theta^{(i)}\}_{i=1}^{t-1}.
\end{equation}
Each entry is a \textit{key-value} pair, where the key $\hat{\mathbf{z}}_{i}\in\mathbb{R}^{d'}$is a low-dimensional fingerprint and the value $\Theta^{(i)}$ is the corresponding checkpoint fine-tuned at time step $i$. To compute the key $\hat{\mathbf{z}}_{i}$, we extract intermediate features from the $i$-th input batch $\mathbf{x}_i$ using a pretrained feature extractor  $\phi_{\Theta^{(0)}}$ and randomly project them to a low-dimensional subspace for efficiency:
\begin{equation}
    \hat{\mathbf{z}}_{i} = \mathrm{RandProj}(\phi_{\Theta^{(0)}}(\mathbf{x}_i)).
\end{equation}
Here, $\mathrm{RandProj}(\cdot):\mathbb{R}^{d}\mapsto\mathbb{R}^{d'}$ is implemented via a fixed Gaussian projection matrix where $d'\ll d$ ensures the keys are compact. As the test-time adaptation progresses, we update the codebook incrementally by appending new pairs, \textit{i.e.,}  $\mathcal{C}^{(t+1)}\leftarrow(\hat{\mathbf{z}_t}, \Theta^{(t)})$. 
\begin{figure*}[t]\vspace{-2ex}
    \includegraphics[width=1\linewidth]{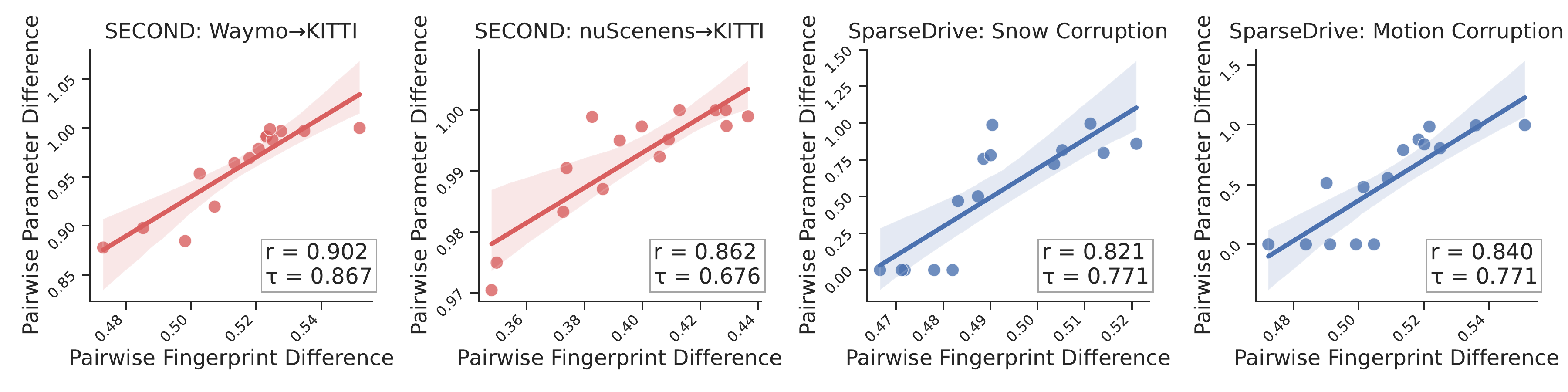}\vspace{-1ex}
    \caption{Pairwise fingerprint differences correlate strongly with model weight differences (Pearson $r$ and  Kendall Tau $\tau$ > 0.7) across SparseDrive \cite{SparseDrive} and SECOND \cite{DBLP:journals/sensors/YanML18}, showing that the low-dimensional fingerprint space reliably reflects parameter space structure.\vspace{-2ex} }\label{fig:corr}
\end{figure*}
\vspace{-1ex}
\subsection{Curvature-Aware Merge Scores}\label{sec:fingerprints}\vspace{-1ex}
To determine which checkpoints in the codebook should be merged at time step $t$, we first compute a merge score for each checkpoint $\Theta^{(i)}\in\mathcal{C}^{(t)}$ using the ridge leverage score. 

     \fcolorbox{gray!30}{gray!10}{\parbox{1\linewidth}{\begin{definition}[Ridge Leverage Score (RLS)]
                    Let $\hat{\mathbf{Z}}_{t-1} = [\hat{\mathbf{z}}_1, \ldots, \hat{\mathbf{z}}_{t-1}]\in\mathbb{R}^{(t-1)\times d'}$ be the matrix of all stored keys (fingerprints), where $\hat{\mathbf{z}_i}$ be the fingerprint of the $i$-th candidate model $\Theta^{(i)}$. We define the ridge leverage scores of the fingerprint $\hat{\mathbf{z}_i}$ as
                    \begin{align*}
                            s_i^{(t)} = \hat{\mathbf{z}}_i^\top \left(\frac{1}{K} \hat{\mathbf{Z}}_{t-1}^\top \hat{\mathbf{Z}}_{t-1} + \lambda I\right)^{-1}\hat{\mathbf{z}}_{i},
                    \end{align*}
                    where $\lambda$ is a regularization parameter. A high leverage score indicates  $\hat{\mathbf{z}}_i$ is influential and lessly observed within the current feature space defined by past direction.   
                \end{definition}}}
                
\noindent \textbf{Theoretical Analysis.} We now connect this leverage score to the inverse of curvature through the lens of LMC.
We begin by revisiting the LMC assumption (Eq. \eqref{eq:lmc}) through a second-order Taylor expansion around $\Theta^{(0)}$:
\begin{equation}
\mathcal{L}(\Theta^{(i)}) \approx \mathcal{L}(\Theta^{(0)}) + \nabla \mathcal{L}^\top\delta_i  + \frac{1}{2} \delta_i^\top H \delta_i, ~\mathrm{with}~ H:=\nabla^2_{\theta}\mathcal{L} (\Theta^{(0)}) ,
\end{equation}
where $\delta_i := \Theta^{(i)}- \Theta^{(0)}$ refers the model update direction and $\mathbf{H}$ is the Hessian at $\Theta^{(0)}$. In this view, the curvature along $\delta_i$ is quantified by the quadratic term $\delta_i^\top H \delta_i$. Its inverse $\delta_i^\top H^{=1} \delta_i$ suggests $\delta_i$ explores a novel region of loss landscape, making it an indicator for selecting diverse checkpoints. 

However, computing the full Hessian in high-dimensional parameter space is impractical, especially in TTA tasks. However, considering that 3D object detection models commonly use linear layers as final regression heads, we can effectively analyze curvature through the simpler and analytically tractable ridge regression setting. Specifically, assume a linear regression head parameterized by weights $w \in \mathbb{R}^{d}$ and a fixed feature extractor $\phi(\cdot)$, yielding a ridge regression objective of the form:\vspace{-1ex}
\begin{align}
\mathcal{L} = \frac{1}{N} \sum_{i=1}^N \| w^\top \phi(\mathbf{x}_i) - \mathbf{y}_i \|^2 + \lambda \|w\|^2,
H_w = 2(\frac{1}{K} \mathbf{Z}^\top \mathbf{Z} + \lambda I),\vspace{-1ex}
\end{align}
where $H_w$ is Hessian matrix in parameter space. More precisely, this reveals the inverse of parameter-space curvature is linked to the proposed ridge leverage score under the low-rank surrogate $\hat{\mathbf{Z}}_{t-1}^\top \hat{\mathbf{Z}}_{t-1}$:\vspace{-1ex}
\begin{equation}
\mathbf{z}_i^\top H_w^{-1} \mathbf{z}_i=  \mathbf{z}_i^\top\left( \frac{2}{K} \mathbf{Z}_{t-1}^\top \mathbf{Z}_{t-1} + 2\lambda I \right)^{-1}\mathbf{z}_i \propto s^{(t)}_i.
\end{equation}
 Empirical analysis (see Fig. \ref{fig:corr}) confirms that fingerprint vectors strongly correlate (Pearson correlation and Kendall Tau scores often exceeding 0.7) with parameter deltas, confirming that the geometry of fingerprint space reliably mirrors that of parameter space.
\vspace{-1ex}

\vspace{-1ex}
\subsection{Model Merging}\label{sec:merge}\vspace{-1ex}
To perform stable model merging, we select top-$K$ high-scoring checkpoints based on ridge leverage scores, yet their associated parameter directions may exhibit destructive interference. To resolve such conflicts, we adopt a sign-consistent merging inspired by ~\citep{DBLP:conf/nips/YadavTCRB23}, which aligns model parameters based on majority sign consensus before merging. Let $\{\Theta^{(i)}\}_{i=1}^{K}$ denote the top-$K$ selected checkpoints and $\{s_i^{(t)}\}_{i=1}^{K}$ their corresponding leverage scores.
For each parameter dimension $j$, we compute the majority sign $\operatorname{sign}_{\mathrm{maj}}(j) := \operatorname{mode}(\{\operatorname{sign}(\Theta_j^{(i)})\}_{i=1}^K)$, and zero out inconsistent components. The merged model is then given by:\vspace{-1ex}
\begin{equation}
    \bar{\Theta}^{(t)} = \sum_{i=1}^{K} \tilde{s}_i^{(t)} \cdot \mathbb{I}\left[\operatorname{sign}(\Theta^{(i)}) = \operatorname{sign}_{\mathrm{maj}}\right] \odot \Theta^{(i)},~ \tilde{s}_i^{(t)} = \frac{s_i^{(t)}}{\sum_{j=1}^{K} s_j^{(t)}}
\end{equation}
where $\odot$ denotes element-wise multiplication, and $\mathbb{I}[\cdot]$ is a binary mask that retains only parameters aligned with the majority sign. This sign-consistent merge ensures coherent parameter updates and stabilizes adaptation under distribution shifts. 

\noindent \textbf{Optimization.} Following the protocol in \cite{DBLP:conf/iclr/ChenMB0HL25}, we use the merged model to generate pseudo-labeled bounding boxes for self-training the LiDAR-based detector online. In realistic end-to-end AD systems (see Fig.~\ref{fig:teaser}), perception, mapping, and planning modules are often integrated into a monolithic architecture. For efficiency, we freeze all components except for the 3D box regression head. Experiments show that CodeMerge not only improves detection performance but also yields gains in downstream mapping and planning \textit{without} requiring additional training or modifications (Table \ref{tab:nuscenes_corruption_mapping_prediction_planning}).

\vspace{-2ex}
\section{Experiments}\vspace{-1ex}
\subsection{Experimental Setup}\vspace{-1ex}

\noindent \textbf{Datasets and Tasks.}  
We conduct comprehensive experiments across five benchmarks for end-to-end autonomous driving and outdoor 3D object detection: \textbf{KITTI}~\citep{DBLP:conf/cvpr/GeigerLU12}, \textbf{KITTI-C}~\citep{DBLP:conf/iccv/KongLLCZRPCL23}, \textbf{Waymo}~\citep{DBLP:conf/cvpr/SunKDCPTGZCCVHN20}, \textbf{nuScenes}~\citep{DBLP:conf/cvpr/CaesarBLVLXKPBB20}, and \textbf{nuScenes-C}~\citep{DBLP:journals/pami/XieKZRPCL25}.  For test-time adaptation in end-to-end autonomous driving, we pre-train models on the nuScenes driving benchmark and adapt them to eight real‐world corruptions in nuScenes-C: Motion Blur (Motion), Color Quantization (Quant), Low Light (Dark), Brightness (Bright), Snow, Fog, Camera Crash (Crash), and Frame Lost.  For LiDAR‐based 3D object detection, we first tackle cross-dataset adaptation (Waymo $\!\rightarrow\!$ KITTI, nuScenes $\!\rightarrow\!$ KITTI) following~\citep{DBLP:conf/cvpr/YangS00Q21,yang2022st3d++,DBLP:conf/iccv/ChenL0BH23}, addressing both object‐level shifts (\textit{e.g.}, scale and point density) and environmental differences (\textit{e.g.}, deployment location, beam configuration). We then evaluate adaptation to sensor failures and weather effects via KITTI $\!\rightarrow\!$ KITTI-C, covering Fog, Wet Conditions (Wet.), Snow, Motion Blur (Moti.), Missing Beams (Beam.), Crosstalk (Cross.T), Incomplete Echoes (Inc.), and Cross-Sensor (Cross.S). The detailed evaluation metric and implementation details can be found in Appendix \ref{sec:implementation}.

\begin{table*}[t]
  \centering
  \caption{\textbf{Perception and tracking results} of the end-to-end SparseDrive model \cite{SparseDrive} with and without TTA on the \textbf{nuScenes-C} \citep{DBLP:journals/pami/XieKZRPCL25} validation set under different corruptions at the highest severity level. The best results for each metric and corruption are highlighted in \textbf{bold}.\vspace{-1ex}}
  \label{tab:nuscenes_perception_tracking_corruptions}
  \renewcommand{\arraystretch}{1.2}
  \Large
  \resizebox{\textwidth}{!}{%
  \begin{tabular}{@{\hspace{\tabcolsep}}lllcccccccccc@{\hspace{\tabcolsep}}}
    \toprule
    \multirow{2}{*}{}      & \multirow{2}{*}{\textsc{\textbf{Corruption}}}      & \multirow{2}{*}{\textsc{\textbf{Method}}}      & \multicolumn{7}{c}{\textsc{\textbf{3D Object Detection}}}      & \multicolumn{3}{c}{\textsc{\textbf{Multi‐Object Tracking}}} \\
    \cmidrule(lr){4-10} \cmidrule(l){11-13}
    & &        & {\textbf{mAP↑}} & \textsc{\textbf{NDS↑}} & \textbf{mATE↓} & \textbf{mASE↓} & \textbf{mAOE↓} & \textbf{mAVE↓} & \textbf{mAAE↓}        & \textbf{AMOTA↑} & \textbf{AMOTP↓} & \textbf{Recall↑} \\
    \midrule

    \multirow{8}{*}{\rotatebox{90}{\textsc{Image Degradation}}}
    & \multirow{4}{*}{\textsc{Motion}}
      & No Adapt. & 0.1468 & 0.3136 & 0.7792 & 0.2908 & 0.8048 & 0.4835 & 0.2398 & 0.0896 & 1.7983 & 0.1837 \\
    & & Tent \citep{DBLP:conf/iclr/WangSLOD21}       & 0.2462   & 0.4113      & 0.6802      & 0.2839      &  \textbf{0.6039}     &  \textbf{0.3243}     & 0.2264      & 0.1736      & \textbf{1.5122}      & 0.2918      \\
    & & MOS \citep{DBLP:conf/iclr/ChenMB0HL25}       & 0.2611      & 0.4125      & 0.6848      & 0.2827      & 0.6588      & 0.3455      & \textbf{0.2087}      & 0.1902      & 1.5239      & 0.3332      \\
    & & \cellcolor{LightGrey} Ours      & \cellcolor{LightGrey}\textbf{0.2759} & \cellcolor{LightGrey}\textbf{0.4206} & \cellcolor{LightGrey}\textbf{0.6697} & \cellcolor{LightGrey}\textbf{0.2815} & \cellcolor{LightGrey}0.6437 & \cellcolor{LightGrey}0.3618 & \cellcolor{LightGrey}0.2169 & \cellcolor{LightGrey}\textbf{0.2192} & \cellcolor{LightGrey}1.5485 & \cellcolor{LightGrey}\textbf{0.3456} \\
    \cmidrule(l){2-13}
    & \multirow{4}{*}{\textsc{Quant}}
      & No Adapt. & 0.2022 & 0.3767 & 0.7095 & 0.2896 & 0.6478 & 0.3814 & 0.2160 & 0.1548 & 1.5398 & 0.2873 \\
    & & Tent \citep{DBLP:conf/iclr/WangSLOD21}       & 0.1424      & 0.3043      & \textbf{0.6527}      & 0.4169      & 0.6032      & 0.5758      & 0.4200      & 0.0981      & 1.6930      & 0.1788      \\
    & & MOS \citep{DBLP:conf/iclr/ChenMB0HL25}       & 0.2560      & 0.4172      & 0.6781      & 0.2848      & 0.6115      & 0.3103     & 0.2231      & 0.2096      & 1.5195      & 0.3287      \\
    & & \cellcolor{LightGrey} Ours      & \cellcolor{LightGrey}\textbf{0.2742} & \cellcolor{LightGrey}\textbf{0.4331} & \cellcolor{LightGrey}0.6575 & \cellcolor{LightGrey}\textbf{0.2764} & \cellcolor{LightGrey}\textbf{0.5903} & \cellcolor{LightGrey}\textbf{0.3018} & \cellcolor{LightGrey}\textbf{0.2137} & \cellcolor{LightGrey}\textbf{0.2339} & \cellcolor{LightGrey}\textbf{1.4868} & \cellcolor{LightGrey}\textbf{0.3330} \\
    \midrule

    \multirow{8}{*}{\rotatebox{90}{\textsc{Illumination Shift}}}
    & \multirow{4}{*}{\textsc{Dark}}
      & No Adapt. & 0.1386 & 0.2804 & 0.7375 & 0.4180 & 0.6880 & 0.6285 & 0.4164 & 0.1169 & 1.7520 & 0.1995 \\
    & & Tent \citep{DBLP:conf/iclr/WangSLOD21}       & 0.1266      & 0.2795      & 0.7243      & 0.4116      & \textbf{0.6396}      & 0.6474      & 0.4151      & 0.0776      & 1.7014      & 0.1697      \\
    & & MOS \citep{DBLP:conf/iclr/ChenMB0HL25}       & 0.1726      & 0.350      & 0.7482      & 0.292      & 0.657      & 0.4202      & 0.2459      & 0.1399      & 1.7148      & 0.2153      \\
    & & \cellcolor{LightGrey} Ours      & \cellcolor{LightGrey}\textbf{0.2060} & \cellcolor{LightGrey}\textbf{0.3727} & \cellcolor{LightGrey}\textbf{0.7206} & \cellcolor{LightGrey}\textbf{0.2852} & \cellcolor{LightGrey}0.6782 & \cellcolor{LightGrey}\textbf{0.3993} & \cellcolor{LightGrey}\textbf{0.2196} & \cellcolor{LightGrey}\textbf{0.1762} & \cellcolor{LightGrey}\textbf{1.6333} & \cellcolor{LightGrey}\textbf{0.2557} \\
    \cmidrule(l){2-13}
    & \multirow{4}{*}{\textsc{Bright}}
      & No Adapt. & 0.3300 & 0.4641 & 0.6355 & \textbf{0.2749} & 0.6084 & 0.3013 & 0.1892 & 0.2829 & 1.4257 & 0.3982 \\
    & & Tent \citep{DBLP:conf/iclr/WangSLOD21}       & 0.2557      & 0.4289      & 0.6345      & 0.2896      & 0.5666      & 0.3143      & \textbf{0.1848}      & 0.1879      & 1.4836      & 0.3002      \\
    & & MOS \citep{DBLP:conf/iclr/ChenMB0HL25}       & 0.3595      & 0.4825      & \textbf{0.6100}      & 0.2757      & 0.6053      & 0.2908      & 0.1909      & 0.3126      & 1.3566      & 0.4387      \\
    & & \cellcolor{LightGrey} Ours      & \cellcolor{LightGrey}\textbf{0.3692} & \cellcolor{LightGrey}\textbf{0.4939} & \cellcolor{LightGrey}0.6138 & \cellcolor{LightGrey}0.2779 & \cellcolor{LightGrey}\textbf{0.5343} & \cellcolor{LightGrey}\textbf{0.2885} & \cellcolor{LightGrey}0.1928 & \cellcolor{LightGrey}\textbf{0.3317} & \cellcolor{LightGrey}\textbf{1.3389} & \cellcolor{LightGrey}\textbf{0.4632} \\
    \midrule

    \multirow{8}{*}{\rotatebox{90}{\textsc{Adverse Weather}}}
    & \multirow{4}{*}{\textsc{Snow}}
      & No Adapt. & 0.0970 & 0.2206 & 0.7974 & 0.4586 & 0.9349 & 0.6614 & 0.4264 & 0.0469 & 1.8822 & 0.1070 \\
    & & Tent \citep{DBLP:conf/iclr/WangSLOD21}       & 0.1417      & 0.2791      & \textbf{0.7312}      & 0.4165      & 0.6904      & 0.6714      & 0.4077      & 0.0779      & 1.7440      & 0.1838      \\
    & & MOS \citep{DBLP:conf/iclr/ChenMB0HL25}       & 0.1478      & 0.3207      & 0.7740      & 0.2995      & 0.7092      & 0.5211      & 0.2284      & 0.0887      & 1.7828      & 0.1747      \\
    & & \cellcolor{LightGrey} Ours      & \cellcolor{LightGrey}\textbf{0.1828} & \cellcolor{LightGrey}\textbf{0.3581} & \cellcolor{LightGrey}0.7558 & \cellcolor{LightGrey}\textbf{0.2930} & \cellcolor{LightGrey}\textbf{0.6009} & \cellcolor{LightGrey}\textbf{0.4604} & \cellcolor{LightGrey}\textbf{0.2222} & \cellcolor{LightGrey}\textbf{0.1136} & \cellcolor{LightGrey}\textbf{1.7119} & \cellcolor{LightGrey}\textbf{0.2293} \\
    \cmidrule(l){2-13}
    & \multirow{4}{*}{\textsc{Fog}} 
      & No Adapt. & 0.3162 & 0.4612 & 0.6295 & 0.2775 & 0.5727 & 0.2984 & \textbf{0.1910} & 0.2756 & 1.4469 & 0.3859 \\
    & & Tent \citep{DBLP:conf/iclr/WangSLOD21}       & 0.2964      & 0.4515      & 0.6372      & 0.2837      & \textbf{0.5190}      & 0.3149      & 0.2121      & 0.2312      & 1.4311      & 0.3623      \\
    & & MOS \citep{DBLP:conf/iclr/ChenMB0HL25}       & 0.3362      & 0.469      & 0.6339      & 0.2797      & 0.5798      & \textbf{0.2961}      & 0.2019      & 0.2907      & 1.3833      & 0.4007      \\
    & & \cellcolor{LightGrey} Ours      & \cellcolor{LightGrey}\textbf{0.3421} & \cellcolor{LightGrey}\textbf{0.4761} & \cellcolor{LightGrey}\textbf{0.6184} & \cellcolor{LightGrey}\textbf{0.2739} & \cellcolor{LightGrey}0.5597 & \cellcolor{LightGrey}0.2995 & \cellcolor{LightGrey}0.1981 & \cellcolor{LightGrey}\textbf{0.2997} & \cellcolor{LightGrey}\textbf{1.3749} & \cellcolor{LightGrey}\textbf{0.4124} \\
    \midrule

    \multirow{8}{*}{\rotatebox{90}{\textsc{Sensor Failures}}}
    & \multirow{4}{*}{\textsc{Crash}}
      & No Adapt. & 0.0785 & 0.2753 & \textbf{0.6467} & 0.4060 & 0.6078 & 0.5953 & 0.3840 & 0.0670 & \textbf{1.8241} & 0.1519 \\
    & & Tent \citep{DBLP:conf/iclr/WangSLOD21}       & 0.0722      & 0.2679      & 0.7426      & 0.3469      & 0.6294      & 0.6658      & 0.2976      & 0.0462      & 1.9007      & 0.1155      \\
    & & MOS \citep{DBLP:conf/iclr/ChenMB0HL25}       & 0.0702 & 0.2659 & 0.7614 & 0.3460 & 0.6169 & 0.6685 & 0.2990 & 0.0454 & 1.8978 & 0.1155 \\
    & & \cellcolor{LightGrey} Ours      & \cellcolor{LightGrey}\textbf{0.0973} & \cellcolor{LightGrey}\textbf{0.3288} & \cellcolor{LightGrey}0.6979 & \cellcolor{LightGrey}\textbf{0.2889} & \cellcolor{LightGrey}\textbf{0.6061} & \cellcolor{LightGrey}\textbf{0.4175} & \cellcolor{LightGrey}\textbf{0.1876} & \cellcolor{LightGrey}\textbf{0.0810} & \cellcolor{LightGrey}1.8372 & \cellcolor{LightGrey}\textbf{0.1550} \\
    \cmidrule(l){2-13}
    & \multirow{4}{*}{\textsc{Lost}}
      & No Adapt. & 0.0886 & 0.3109 & \textbf{0.7314} & 0.2792 & 0.6206 & 0.4717 & 0.2310 & 0.0549 & 1.7638 & 0.1644 \\
    & & Tent \citep{DBLP:conf/iclr/WangSLOD21}       & 0.0372      & 0.2371      & 0.8386      & 0.2913      & 0.7439      & 0.7068      & 0.2337      & 0.0029      & 1.9856      & 0.0406      \\
    & & MOS \citep{DBLP:conf/iclr/ChenMB0HL25}       & 0.0479      & 0.2116      & 0.8913      & 0.3464      & 0.7567      & 0.8008      & 0.3281      & 0.0131      & 1.9670      & 0.0624      \\
    & & \cellcolor{LightGrey} Ours      & \cellcolor{LightGrey}\textbf{0.1172} & \cellcolor{LightGrey}\textbf{0.3292} & \cellcolor{LightGrey}0.7638 & \cellcolor{LightGrey}\textbf{0.2787} & \cellcolor{LightGrey}\textbf{0.5810} & \cellcolor{LightGrey}\textbf{0.4461} & \cellcolor{LightGrey}\textbf{0.2243} & \cellcolor{LightGrey}\textbf{0.0700} & \cellcolor{LightGrey}\textbf{1.7605} & \cellcolor{LightGrey}\textbf{0.1788} \\
    \midrule

    \multicolumn{2}{c}{\multirow{4}{*}{\textbf{\textsc{Average}}}}
      & No Adapt. & 0.1747 & 0.3378 & 0.7083 & 0.3368 & 0.6856 & 0.4777 & 0.2867 & 0.1361 & 1.6791 & 0.2347 \\
    \multicolumn{2}{c}{}
      & Tent \citep{DBLP:conf/iclr/WangSLOD21} 
      & 0.1648 & 0.3325 & 0.7052 & 0.3426 & 0.6245 & 0.5276 & 0.2997 & 0.1119 & 1.6815 & 0.2053 \\
    \multicolumn{2}{c}{}
      & MOS \citep{DBLP:conf/iclr/ChenMB0HL25}       
      & 0.2028 & 0.3551 & 0.7269 & 0.3205 & 0.6633 & 0.4829 & 0.2711 & 0.1599 & 1.6461 & 0.2532 \\
    \multicolumn{2}{c}{}
      & \cellcolor{LightGrey}\textbf{Ours}
      & \cellcolor{LightGrey}\textbf{0.2331} & \cellcolor{LightGrey}\textbf{0.4016} & \cellcolor{LightGrey}\textbf{0.6872} & \cellcolor{LightGrey}\textbf{0.2819} 
      & \cellcolor{LightGrey}\textbf{0.5993} & \cellcolor{LightGrey}\textbf{0.3719} & \cellcolor{LightGrey}\textbf{0.2094} & \cellcolor{LightGrey}\textbf{0.1907} 
      & \cellcolor{LightGrey}\textbf{	1.5865} & \cellcolor{LightGrey}\textbf{0.2966} \\
    \bottomrule
  \end{tabular}
  }\vspace{-2ex}
\end{table*}

\begin{table*}[t]
  \centering
  \caption{\textbf{Impact of TTA on downstream modules of end-to-end SparseDrive~\cite{SparseDrive}.} We evaluate online mapping, motion prediction, and trajectory planning on the \textbf{nuScenes-C}~\citep{DBLP:journals/pami/XieKZRPCL25} under the highest severity of various corruptions. These modules are not fine-tuned; all performance gains stem from TTA applied to the detection module. Best results per metric and corruption are shown in \textbf{bold}.\vspace{-1ex}}
  \label{tab:nuscenes_corruption_mapping_prediction_planning}
  \renewcommand{\arraystretch}{1.2}
  \Large
  \resizebox{\textwidth}{!}{%
  \begin{tabular}{@{\hspace{\tabcolsep}}lllcccccccccc@{\hspace{\tabcolsep}}}
    \toprule
    \multirow{2}{*}{} & \multirow{2}{*}{\textsc{\textbf{Corruption}}} & \multirow{2}{*}{\textsc{\textbf{Method}}}
      & \multicolumn{4}{c}{\textsc{\textbf{Online Mapping}}}
      & \multicolumn{4}{c}{\textsc{\textbf{Motion Prediction}}}
      & \multicolumn{2}{c}{\textsc{\textbf{Planning}}} \\

    \cmidrule(lr){4-7} \cmidrule(lr){8-11} \cmidrule(l){12-13}
    & & &   \textbf{AP$_{\text{ped}}$↑} & \textbf{AP$_d$↑} & \textbf{AP$_b$↑} &\textbf{mAP↑}
            & \textbf{mADE↓} & \textbf{mFDE↓} & \textbf{MR↓} & \textbf{EPA↑}
            & \textbf{L2-Avg↓} & \textbf{CR-Avg↓} \\
    \midrule

% =========================  Image Degradation  =========================
    \multirow{8}{*}{\rotatebox{90}{\textsc{Image Degradation}}}
    & \multirow{4}{*}{\textsc{Motion}}
        & No Adapt. & 0.1988 & 0.2343 & 0.1999 & 0.2110 & 0.8630 & 1.3483 & 0.1750 & 0.2616 & 0.7877 & 0.215 \\
    &   & Tent \citep{DBLP:conf/iclr/WangSLOD21}       & 0.3425      & 0.3794      & 0.3876      & 0.3698      & 0.7786      & 1.1825      & 0.1520      & 0.3712      & \textbf{0.6474}      & \textbf{0.090}      \\
    &   & MOS \citep{DBLP:conf/iclr/ChenMB0HL25}      & 0.3452      & 0.3943      & 0.4012      & 0.3802      & 0.7348      & 1.1278      & \textbf{0.1560}      & 0.3742      & 0.6694      & 0.134      \\
    &   & \cellcolor{LightGrey}\textbf{Ours}
                     & \cellcolor{LightGrey}\textbf{0.3660} & \cellcolor{LightGrey}\textbf{0.4212} & \cellcolor{LightGrey}\textbf{0.4283} & \cellcolor{LightGrey}\textbf{0.4052}
                     & \cellcolor{LightGrey}\textbf{0.7264} & \cellcolor{LightGrey}\textbf{1.1200} & \cellcolor{LightGrey}0.1570 & \cellcolor{LightGrey}\textbf{0.3945}
                     & \cellcolor{LightGrey}0.6580 & \cellcolor{LightGrey}0.110 \\

    \cmidrule(l){2-13}
    & \multirow{4}{*}{\textsc{Quant}}
        & No Adapt. & 0.1742 & 0.2317 & 0.2069 & 0.2043 & 0.7620 & 1.1734 & 0.1526 & 0.3204 & 0.7301 & 0.159 \\
    &   & Tent \citep{DBLP:conf/iclr/WangSLOD21}       &  0.1526      & 0.2153      & 0.2088 &0.1922          & 0.8489      & 1.3551      & 0.1602      & 0.2987      & 0.6966      & 0.120      \\
    &   & MOS \citep{DBLP:conf/iclr/ChenMB0HL25}             & 0.2346      & 0.3208      & 0.2918  & 0.2824    & 0.7040      & \textbf{1.0822}      & \textbf{0.1445}      & 0.3668      & 0.6848      & \textbf{0.118}      \\
    &   & \cellcolor{LightGrey}\textbf{Ours}
                      & \cellcolor{LightGrey}\textbf{0.2600} & \cellcolor{LightGrey}\textbf{0.3445} & \cellcolor{LightGrey}\textbf{0.3267}& \cellcolor{LightGrey}\textbf{0.3104}
                     & \cellcolor{LightGrey}\textbf{0.7002} & \cellcolor{LightGrey}1.0859 & \cellcolor{LightGrey}0.1454 & \cellcolor{LightGrey}\textbf{0.3840}
                     & \cellcolor{LightGrey}\textbf{0.6762} & \cellcolor{LightGrey}0.125 \\

% =========================  Illumination Shift  =========================
    \midrule
    \multirow{8}{*}{\rotatebox{90}{\textsc{Illumination Shift}}}
    & \multirow{4}{*}{\textsc{Dark}}
        & No Adapt. & 0.1173 & 0.2038 & 0.1812 & 0.1675 & 0.8428 & 1.3255 & 0.1714 & 0.2757 & 0.7535 & 0.276 \\
    &   & Tent \citep{DBLP:conf/iclr/WangSLOD21}             & 0.2116      & 0.256      & 0.2481 & 0.2386     & 0.8603      & 1.3314      & 0.1786      & 0.2722      & 0.7049      & 0.123      \\
    &   & MOS \citep{DBLP:conf/iclr/ChenMB0HL25}             & 0.2261      & 0.3090      & 0.2892   & 0.2748   & 0.7956      & 1.2443      & 0.1730      & 0.3066      & 0.6824      & 0.136      \\
    &   & \cellcolor{LightGrey}\textbf{Ours}
                      & \cellcolor{LightGrey}\textbf{0.2825} & \cellcolor{LightGrey}\textbf{0.3637} & \cellcolor{LightGrey}\textbf{0.3291}
                      & \cellcolor{LightGrey}\textbf{0.3251}
                     & \cellcolor{LightGrey}\textbf{0.7493} & \cellcolor{LightGrey}\textbf{1.1639} & \cellcolor{LightGrey}\textbf{0.1644} & \cellcolor{LightGrey}\textbf{0.3397}
                     & \cellcolor{LightGrey}\textbf{0.6602} & \cellcolor{LightGrey}\textbf{0.117} \\

    \cmidrule(l){2-13}
    & \multirow{4}{*}{\textsc{Bright}}
        & No Adapt. & 0.3777 & 0.4847 & 0.4833 & 0.4486 & 0.6646 & 1.0246 & 0.1369 & 0.4468 & 0.6306 & 0.126 \\
    &   & Tent \citep{DBLP:conf/iclr/WangSLOD21}            & 0.3550      & 0.4342      & 0.4591 & 0.4161     & 0.6882      & 1.0739      & 0.1369      & 0.3978      & 0.6487      & 0.095      \\
    &   & MOS \citep{DBLP:conf/iclr/ChenMB0HL25}            & 0.4053      & 0.4960      & 0.5127  & 0.4713     & \textbf{0.6468}      & \textbf{1.0031}      & \textbf{0.1357}      & 0.4593      & 0.6243      & 0.123      \\
    &   & \cellcolor{LightGrey}\textbf{Ours}
                      & \cellcolor{LightGrey}\textbf{0.4305} & \cellcolor{LightGrey}\textbf{0.5224} & \cellcolor{LightGrey}\textbf{0.5398}& \cellcolor{LightGrey}\textbf{0.4976}
                     & \cellcolor{LightGrey}0.6504 & \cellcolor{LightGrey}1.0122 & \cellcolor{LightGrey}0.1392 & \cellcolor{LightGrey}\textbf{0.468}
                     & \cellcolor{LightGrey}\textbf{0.6209} & \cellcolor{LightGrey}\textbf{0.094} \\

% =========================  Adverse Weather  ============================
    \midrule
    \multirow{8}{*}{\rotatebox{90}{\textsc{Adverse Weather}}}
    & \multirow{4}{*}{\textsc{Snow}}
        & No Adapt. & 0.0061 & 0.0322 & 0.0369 & 0.0250 & 1.0643 & 1.7042 & 0.1930 & 0.2113 & 0.8897 & 0.431 \\
    &   & Tent \citep{DBLP:conf/iclr/WangSLOD21}             & 0.1083      & 0.1320      & 0.1359  & 0.1254    & 0.9147      & 1.4192      & 0.1753      & 0.2804      & \textbf{0.7552}      & \textbf{0.132}      \\
    &   & MOS \citep{DBLP:conf/iclr/ChenMB0HL25}             & \textbf{0.1237}      & 0.1564      & 0.1545   & 0.1448   & 0.8736      & 1.3476      & 0.1737      & 0.2994      & 0.7684      & 0.192     \\
    &   & \cellcolor{LightGrey}\textbf{Ours}
                     & \cellcolor{LightGrey}0.1134 & \cellcolor{LightGrey}\textbf{0.1812} & \cellcolor{LightGrey}\textbf{0.1740} & \cellcolor{LightGrey}\textbf{0.1562}
                     & \cellcolor{LightGrey}\textbf{0.8074} & \cellcolor{LightGrey}\textbf{1.2589} & \cellcolor{LightGrey}\textbf{0.1717} & \cellcolor{LightGrey}\textbf{0.3135}
                     & \cellcolor{LightGrey}0.7634 & \cellcolor{LightGrey}0.190 \\

    \cmidrule(l){2-13}
    & \multirow{4}{*}{\textsc{Fog}}
        & No Adapt. & 0.3600 & 0.4649 & 0.4076 & 0.4109 & \textbf{0.6482} & \textbf{0.9904} & \textbf{0.1347} & 0.4380 & 0.6257 & 0.105 \\
    &   & Tent \citep{DBLP:conf/iclr/WangSLOD21}           & 0.3786      & 0.4492      & 0.4438   & 0.4239     & 0.6861      & 1.0631      & 0.1405      & 0.4182      & 0.6533      & \textbf{0.087}      \\
    &   & MOS \citep{DBLP:conf/iclr/ChenMB0HL25}             & 0.4161      & 0.4950      & 0.4785  & 0.4632    & 0.6549      & 1.0087      & 0.1401      & 0.4539      & 0.6225      & 0.106      \\
    &   & \cellcolor{LightGrey}\textbf{Ours}
                     & \cellcolor{LightGrey}\textbf{0.4276} & \cellcolor{LightGrey}\textbf{0.5022} & \cellcolor{LightGrey}\textbf{0.4843} & \cellcolor{LightGrey}\textbf{0.4714}
                     & \cellcolor{LightGrey}0.6501 & \cellcolor{LightGrey}1.0008 & \cellcolor{LightGrey}0.1394 & \cellcolor{LightGrey}\textbf{0.4557}
                     & \cellcolor{LightGrey}\textbf{0.6200} & 0.110 \\

% =========================  Sensor Failures  ============================
    \midrule
    \multirow{8}{*}{\rotatebox{90}{\textsc{Sensor Failures}}}
    & \multirow{4}{*}{\textsc{Crash}}
        & No Adapt. & \textbf{0.1029} & 0.1019 & \textbf{0.0618} & \textbf{0.0889} & 0.8662 & 1.3375 & 0.1652 & 0.1920 & 0.9276 & \textbf{0.374} \\
    &   & Tent \citep{DBLP:conf/iclr/WangSLOD21}             & 0.0431      & 0.0764      & 0.0141   & 0.0445   & 0.8691      & 1.3548      & 0.1710      & 0.1771      & 0.8852      & 0.704      \\
    &   & MOS \citep{DBLP:conf/iclr/ChenMB0HL25}       & 0.0394      & 0.0706      & 0.0100      & 0.0400      & 0.8878      & 1.3895      & 0.1766     & 0.1721      & 0.8977      & 0.730      \\
    &   & \cellcolor{LightGrey}\textbf{Ours}
                      & \cellcolor{LightGrey}0.0727 & \cellcolor{LightGrey}\textbf{0.1154} & \cellcolor{LightGrey}0.0279& \cellcolor{LightGrey}0.0720
                     & \cellcolor{LightGrey}\textbf{0.8302} & \cellcolor{LightGrey}\textbf{1.3022} & \cellcolor{LightGrey}\textbf{0.1637} & \cellcolor{LightGrey}\textbf{0.1974}
                     & \cellcolor{LightGrey}\textbf{0.8539} & \cellcolor{LightGrey}0.630 \\

    \cmidrule(l){2-13}
    & \multirow{4}{*}{\textsc{Lost}}
        & No Adapt. & 0.0892 & 0.0388 & \textbf{0.0250} & 0.0510 & 1.0327 & 1.4772 & 0.1740 & \textbf{0.1826} & 0.9932 & \textbf{0.483} \\
    &   & Tent \citep{DBLP:conf/iclr/WangSLOD21}            & 0.0431      & \textbf{0.0547}     & 0.0163   & 0.0380    & 1.4194      & 2.1114      & 0.2383      & 0.0737      & 0.9985     & 0.622      \\
    &   & MOS \citep{DBLP:conf/iclr/ChenMB0HL25}             & 0.0180      & 0.0153      & 0.0038 &  0.0124    & 1.5468      & 2.3163      & 0.2155      & 0.0873      & 1.0628      & 0.734      \\  & &\cellcolor{LightGrey}\textbf{Ours}
                      & \cellcolor{LightGrey}\textbf{0.0723} & \cellcolor{LightGrey}0.0503 & \cellcolor{LightGrey}\textbf{0.0250} & \cellcolor{LightGrey}\textbf{0.0492}
                     & \cellcolor{LightGrey}\textbf{1.0004} & \cellcolor{LightGrey}\textbf{1.4304} & \cellcolor{LightGrey}\textbf{0.1739} & \cellcolor{LightGrey}0.0952
                     & \cellcolor{LightGrey}\textbf{0.9600} & \cellcolor{LightGrey}0.661 \\

% =========================  Average Results  ============================

    \midrule
    \multicolumn{2}{c}{\multirow{4}{*}{\textbf{\textsc{Average}}}}
      & No Adapt.      
      & 0.1783 & 0.2240 & 0.2003 & 0.2009
      & 0.8430 & 1.2976 & 0.1629 & 0.2911
      & 0.7923 & 0.2711 \\
    \multicolumn{2}{c}{}  
      & Tent \citep{DBLP:conf/iclr/WangSLOD21}
      & 0.2044 & 0.2497 & 0.2392 & 0.2311
      & 0.8832 & 1.3614 & 0.1691 & 0.2862
      & 0.7487 & \textbf{0.2466} \\
    \multicolumn{2}{c}{}
      & MOS       
      & 0.2260 & 0.2822 & 0.2677 & 0.2586
      & 0.8555 & 1.3149 & 0.1644 & 0.3150
      & 0.7515 & 0.2841 \\
    \multicolumn{2}{c}{}
      & \cellcolor{LightGrey}\textbf{Ours}      
      & \cellcolor{LightGrey}\textbf{0.2531} 
      & \cellcolor{LightGrey}\textbf{0.3126} 
      & \cellcolor{LightGrey}\textbf{0.2919} 
      & \cellcolor{LightGrey}\textbf{0.2859}
      & \cellcolor{LightGrey}\textbf{0.7643} 
      & \cellcolor{LightGrey}\textbf{1.1718} 
      & \cellcolor{LightGrey}\textbf{0.1568} 
      & \cellcolor{LightGrey}\textbf{0.3312}
      & \cellcolor{LightGrey}\textbf{0.7266} 
      & \cellcolor{LightGrey}0.2546 \\
    \bottomrule
  \end{tabular}%
  }\vspace{-2ex}
\end{table*}

\noindent\textbf{Baselines.} 
We compare the proposed CodeMerge against a broad range of methods:  
(i) \textbf{No Adapt.},  the pretrained model evaluated directly on the target datasets;  
(ii) \textbf{SN}~\citep{DBLP:conf/cvpr/WangCYLHCWC20}, a \textit{weakly supervised DA} technique that rescales source objects using target size statistics;  
(iii) \textbf{ST3D}~\citep{DBLP:conf/cvpr/YangS00Q21}, the first \textit{UDA} method for 3D detection, employing multi‐epoch self‐training with pseudo labels; 
(iv) \textbf{Tent}~\citep{DBLP:conf/iclr/WangSLOD21}, an \textit{TTA} approach that minimizes prediction entropy;  
(v) \textbf{CoTTA}~\citep{DBLP:conf/cvpr/0013FGD22}, which combines mean‐teacher supervision with stochastic augmentations for \textit{TTA}; 
(vi) \textbf{SAR}~\citep{DBLP:conf/iclr/Niu00WCZT23}, enhancing Tent by sharpness‐aware and reliability‐aware entropy minimization;  
(vii) \textbf{MemCLR}~\citep{DBLP:conf/wacv/VSOP23}, the first \textit{online TTA} method that uses memory‐augmented mean‐teacher for 2D detection;  
(viii) \textbf{Reg-TTA3D}~\citep{DBLP:conf/eccv/YuanZGYSQC24}, which regularizes 3D box regression by enforcing noise‐consistent pseudo labels during \textit{3D TTA};  
(ix) \textbf{MOS}~\citep{DBLP:conf/iclr/ChenMB0HL25}, dynamically fusing a bank of top‐$K$ checkpoints through kernel‐based synergy for \textit{3D TTA};  
(x) \textbf{DPO}~\citep{DBLP:conf/mm/ChenWL0H24}, flattening the test‐time loss landscape via dual perturbations for \textit{3D TTA}.  (xi) \textbf{Oracle}, a \textit{fully supervised} model trained with annotated target datasets.

\vspace{-1ex}
\subsection{Main Results and Analysis}\vspace{-1ex}

\noindent \textbf{TTA on End-to-End Autonomous Driving. }
We comprehensively evaluate our {CodeMerge} method on nuScenes-C~\citep{DBLP:journals/pami/XieKZRPCL25} with the end-to-end SparseDrive model~\cite{SparseDrive}, covering five downstream tasks: 3D detection, multi-object tracking, online mapping, motion prediction, and trajectory planning under diverse corruptions. Table~\ref{tab:nuscenes_perception_tracking_corruptions} shows {CodeMerge} consistently outperforms all baselines, including No Adapt, Tent, and the state-of-the-art MOS~\cite{DBLP:conf/iclr/ChenMB0HL25} in averaged results.  In 3D detection, we boost mAP by \textbf{33.6\%} over no adaptation (0.1747 → 0.2334) and by 13.3\% over MOS. CodeMerge also reduces mASE by 4.4\% relative to MOS, and lower mAVE by 19\%. Under the \textit{Bright} corruption, CodeMerge improves mAP by 11.9\% over no adaptation, with consistent gains in other metrics. In multi-object tracking, {CodeMerge} improves AMOTA by 19.3\%, reduces AMOTP by 13.8\%, and raises recall by 16.5\% when compared with the SOTA baseline, MOS. Notably, under the most safety-critical \textit{Lost} scenario, the proposed method achieves the highest recall (0.1788) and lowest tracking error among all methods. Although only perception weights are adapted, downstream tasks benefit markedly. As reported in Table \ref{tab:nuscenes_corruption_mapping_prediction_planning}. CodeMerge increases online mapping mAP by \textbf{42.3\%} (0.2009 → 0.2859) over no adaptation, with \textbf{+45.7\%} on lane boundaries and +39.5\% on obstacles, especially \textbf{+94.2\%} under \textit{Dark}. For motion prediction, mADE and mFDE fall by 9.3\% and 9.7\% compared to no adaptation, respectively, while EPA (higher is better) rises by 13.8\%. For planning, average lateral deviation falls 8.3\% (0.7923 m → 0.7266m) and collision risk drops 6.1\% compared to no adaptation. These consistent gains achieved without touching non-perception modules, confirm that the proposed lightweight, fingerprint-guided merging framework stabilizes the detector and unlocks robust performance across all autonomous driving tasks.

\begin{table}[t]
  \centering\vspace{-2ex}
  \footnotesize
  \caption{ \textbf{TTA results for LiDAR-based 3D detection across different datasets}.  
  We report $\text{AP}_{\text{BEV}}$ / $\text{AP}_{\text{3D}}$ (moderate).  
  “Oracle’’ = fully–supervised on target;  
  \textbf{Bold} = best; \underline{underline} = second best.}
  \label{tab:cross_dataset_adapt}
  \renewcommand{\arraystretch}{1}
  \resizebox{.95\linewidth}{!}{%
  \begin{tabular}{@{\hspace{\tabcolsep}}lcccccc@{\hspace{\tabcolsep}}}
    \toprule
    \multirow{2}{*}{\textsc{\textbf{Method}}}
        & \multirow{2}{*}{\textsc{\textbf{Venue}}}
        & \multirow{2}{*}{\textsc{\textbf{TTA}}}
        & \multicolumn{2}{c}{\textsc{\textbf{Waymo $\rightarrow$ KITTI}}}
        & \multicolumn{2}{c}{\textsc{\textbf{nuScenes $\rightarrow$ KITTI}}} \\
    \cmidrule(lr){4-5}\cmidrule(l){6-7}
        &   &   & \textbf{AP$_\text{BEV}$ / AP$_\text{3D}$}
                       & \textbf{Closed Gap}
                       & \textbf{AP$_\text{BEV}$ / AP$_\text{3D}$}
                       & \textbf{Closed Gap} \\
    \midrule
    No Adapt.   & –           & \multirow{4}{*}{$\times$} &
                 67.64 / 27.48 & –                     &
                 51.84 / 17.92 & – \\[0.15em]
    SN \citep{DBLP:conf/cvpr/WangCYLHCWC20}          & CVPR’20     & &
                 78.96 / 59.20 & +72.33\% / +69.00\%   &
                 40.03 / 21.23 & +37.55\% / +5.96\% \\[0.15em]
    ST3D \citep{DBLP:conf/cvpr/YangS00Q21}        & CVPR’21     & &
                 82.19 / 61.83 & +92.97\% / +74.72\%   &
                 75.94 / 54.13 & +76.63\% / +65.21\% \\[0.15em]
    Oracle      & –           & &
                 83.29 / 73.45 & –                     &
                 83.29 / 73.45 & – \\
    \midrule
    Tent \citep{DBLP:conf/iclr/WangSLOD21}
                & ICLR’21     & \multirow{8}{*}{$\checkmark$} &
                 65.09 / 30.12 & –16.29\% / +5.74\%  &
                 46.90 / 18.83 & –15.71\% / +1.64\% \\[0.15em]
    CoTTA \citep{DBLP:conf/cvpr/0013FGD22}      & CVPR’22     & &
                 67.46 / 35.34 & –1.15\% / +17.10\%  &
                 68.81 / 47.61 & +53.96\% / +53.47\% \\[0.15em]
    SAR \citep{DBLP:conf/iclr/Niu00WCZT23}        & ICLR’23     & &
                 65.81 / 30.39 & –11.69\% / +6.33\%  &
                 61.34 / 35.74 & +30.21\% / +32.09\% \\[0.15em]
    MemCLR \citep{DBLP:conf/wacv/VSOP23}     & WACV’23     & &
                 65.61 / 29.83 & –12.97\% / +5.11\%  &
                 61.47 / 35.76 & +30.62\% / +32.13\% \\[0.15em]
    DPO  \citep{DBLP:conf/mm/ChenWL0H24}       & MM’24       & &
                 75.81 / 55.74 & +52.20\% / +61.47\% &
                 \underline{73.27} / \underline{54.38} &
                 \underline{+68.13\%} / \underline{+65.66\%} \\[0.15em]
    Reg-TTA3D \citep{DBLP:conf/eccv/YuanZGYSQC24}     & ECCV’24     & &
                 81.60 / 56.03 & +89.20\% / +62.11\% &
                 68.73 / 44.56 & +53.70\% / +47.97\% \\[0.15em]
    MOS \citep{DBLP:conf/iclr/ChenMB0HL25}
                & ICLR’25     & &
                 \underline{81.90} / \underline{64.16} &
                 \underline{+91.12\%} / \underline{+79.79\%} &
                 71.13 / 51.11 & +61.33\% / +59.78\% \\
    \rowcolor{LightGrey}
    \textbf{Ours} & \textbf{–} & &
                 \cellcolor{LightGrey}\textbf{84.62 / 66.31} &
                 \cellcolor{LightGrey}\textbf{+108.5\% / +84.47\%} &
                 \cellcolor{LightGrey}\textbf{77.41 / 58.54} &
                 \cellcolor{LightGrey}\textbf{+81.30\% / +73.15\%} \\
    \bottomrule
  \end{tabular}}\vspace{-2ex}
\end{table}

\noindent\textbf{TTA on LiDAR-based Detection.}  
We examine CodeMerge’s performance in 3D object detection\begin{wraptable}{r}{0.6\textwidth}\vspace{-2ex}
  \centering
  \caption{\textbf{TTA results on KITTI-C.} We evaluate the LiDAR-based SECOND detector \cite{DBLP:journals/sensors/YanML18} under the highest severity level of various corruptions, reporting $\text{AP}_{\text{3D}}$ (hard).}
  \label{tab:kitti-c_corruption_hard}
  \renewcommand{\arraystretch}{1}
  \Large
  \resizebox{\linewidth}{!}{%
  \begin{tabular}{@{\hspace{\tabcolsep}}lcccccccc@{\hspace{\tabcolsep}}}
    \toprule 
        & No Adapt. 
        & Tent \citep{DBLP:conf/iclr/WangSLOD21} 
        & CoTTA \cite{DBLP:conf/cvpr/0013FGD22}
        & SAR \citep{DBLP:conf/iclr/Niu00WCZT23} 
        & MemCLR \citep{DBLP:conf/wacv/VSOP23} 
        & DPO \cite{DBLP:conf/mm/ChenWL0H24}
        & MOS \citep{DBLP:conf/iclr/ChenMB0HL25} 
        & \cellcolor{LightGrey}\textbf{Ours} \\
    \midrule
    Fog      & 68.23 & 68.73 & 68.49 & 68.14 & 68.23 & 68.72 & 69.11 & \cellcolor{LightGrey}\textbf{75.96} \\
    Snow     & 59.07 & 59.50 & 59.45 & 58.78 & 58.74 & 60.80 & 62.72 & \cellcolor{LightGrey}\textbf{63.53} \\
    Inc.     & 25.68 & 26.44 & 27.85 & 26.42 & 27.47 & 27.16 & \textbf{34.53} & \cellcolor{LightGrey}32.18 \\
    CrossT.  & 75.49 & 74.67 & 72.22 & 74.51 & 74.25 & 75.52 & 75.47 & \cellcolor{LightGrey}\textbf{75.76} \\
    Moti.    & 38.21 & 38.15 & 38.62 & 38.12 & 37.57 & 38.71 & 40.59 & \cellcolor{LightGrey}\textbf{44.87} \\
    CrossS.  & 41.08 & 41.17 & 40.80 & 40.63 & 40.90 & 42.09 & \textbf{43.68} & \cellcolor{LightGrey}42.36 \\
    Wet.     & 76.25 & 76.36 & 76.43 & 76.23 & 76.25 & 76.89 & 77.79 & \cellcolor{LightGrey}\textbf{79.82} \\
    Beam.    & 53.93 & 53.85 & 53.98 & 53.75 & 53.49 & 54.06 & 55.91 & \cellcolor{LightGrey}\textbf{57.26} \\
    \midrule
    \textbf{Mean} 
             & 54.74 & 54.86 & 54.73 & 54.57 & 54.61 & 55.49 & 57.48 & \cellcolor{LightGrey} \textbf{58.97} \\
    \bottomrule
  \end{tabular}} \vspace{-2ex}
\end{wraptable} across two distinct types of domain shifts: Cross-dataset (Waymo → KITTI, nuScenes → KITTI) and Corruption-induced shifts (KITTI → KITTI-C). {\textbf{(1) Cross-dataset} (Table~\ref{tab:cross_dataset_adapt}).}  Compared with the non-adapted model, {CodeMerge} lifts AP$_\text{BEV}$ by 25.1\% and AP$_\text{3D}$ by 141\% on Waymo $\!\rightarrow\!$ KITTI, closing 108.5\%/84.5\% of the domain gap and even surpassing the multi-epoch ST3D and fully supervised Oracle in AP$_\text{BEV}$. On nuScenes → KITTI, it narrows the gap by 81.3\%/73.15\%, again outperforming the strongest TTA baselines (MOS, DPO) and exceeding ST3D by +1.9\% AP$_\text{BEV}$ and +8.1\% AP$_\text{3D}$. {\textbf{(2) Corruption-induced} (Table~\ref{tab:kitti-c_corruption_hard}).}  Against KITTI → KITTI-C corruptions, {CodeMerge} raises mean AP$_\text{3D}$ by +7.7\% over no adaptation and +2.6\% over the best prior TTA baseline. Under \emph{Fog} and  \emph{Wet} corruption, gains are pronounced: +9.9\% (75.96 vs.\ 69.11) and +2.6\% (79.82 vs.\ 77.79), respectively, indicating enhanced resilience to visibility and environment degradations. These results demonstrate that our latent-space, fingerprint-guided merging not only closes cross-domain gaps more effectively than existing TTA methods but also surpasses dedicated domain adaptation approaches, providing robust performance across diverse and challenging environments.

\begin{table*}[!tbp]
  \centering
  \caption{Ablation study on different checkpoint selection strategies, number of checkpoints to merge ($K$), and random projection dimension ($d'$) on \textbf{nuScenes-C} \citep{DBLP:journals/pami/XieKZRPCL25} (motion blur at the heaviest level). }
  \label{tab:ablation_nuscenes_c}
  \renewcommand{\arraystretch}{1.2}\vspace{-1ex}
  \Large
  \resizebox{\textwidth}{!}{%
  \begin{tabular}{@{\hspace{\tabcolsep}}lcc*{10}{c}@{\hspace{\tabcolsep}}}
    \toprule
    \multirow{2}{*}{\textsc{\textbf{Merge}}}
      & \multirow{2}{*}{\textsc{\textbf{K}}}
      & \multirow{2}{*}{\textsc{\textbf{Proj.-D}}}
      & \multicolumn{2}{c}{\textsc{\textbf{Detection}}}
      & \multicolumn{2}{c}{\textsc{\textbf{Tracking}}}
      & \multicolumn{2}{c}{\textsc{\textbf{Mapping}}}
      & \multicolumn{2}{c}{\textsc{\textbf{Motion}}}
      & \multicolumn{2}{c}{\textsc{\textbf{Planning}}} \\
    \cmidrule(lr){4-5} \cmidrule(lr){6-7} \cmidrule(lr){8-9}
    \cmidrule(lr){10-11} \cmidrule(l){12-13}
    & & 
      & \textbf{mAP↑} & \textbf{NDS↑}
      & \textbf{AMOTA↑} & \textbf{AMOTP↓}
      & \textbf{mAP↑} & \textbf{AP$_\mathrm{ped}$↑}
      & \textbf{mADE↓} & \textbf{mFDE↓}
      & \textbf{L2-Avg↓} & \textbf{CR-Avg↓} \\
    \midrule
    \rowcolor{LightGrey} Random & 5 & –    & 0.2740 & 0.4185 & 0.2152 & 1.5461 & 0.4011 & 0.3678 & 0.7251 & 1.1192 & 0.6631 & 0.112   \\
    Recent & 5 & –    & 0.2480 & 0.3985 & 0.1866 & 1.6040 & 0.3748
 & 0.3410 & 0.7368 & 1.1436  & 0.6795 & 0.149  \\
    \rowcolor{LightGrey} KMeans++           & 5 & 1024 & 0.2746 & 0.4192 & 0.2157 & 1.5490 & 0.4010 & 0.3678 & 0.7246 & 1.1182 & 0.6625 & 0.105  \\
      Leverage           & 5 & 1024 & \textbf{0.2851} & \textbf{0.4264}   & \textbf{0.2241} & \textbf{1.5206}   & 0.4103 & 0.3713  & 0.7228 & 1.1146   & \textbf{0.6504} & 0.109    \\
    \midrule
    \rowcolor{LightGrey}Leverage           & 3 & 1024 & 0.2655 & 0.4122 & 0.2077 & 1.5630 & 0.3623 & \textbf{0.3928} & 0.7407 & 1.1461 & 0.6651 & 0.120  \\
   
    Leverage           & 9 & 1024 & 0.2818 & 0.4231 & 0.2195 & 1.5240 & \textbf{0.4167} & 0.3814 & \textbf{0.7180} & \textbf{1.1066} & 0.6534 &  0.103 \\
    \midrule
    \rowcolor{LightGrey} Leverage           & 5 & 256  & 0.2749 & 0.4176 & 0.2168 & 1.5488 & 0.4010 & 0.3678 & 0.7228 & 1.1142 & 0.6615 & \textbf{0.096}  \\
    Leverage           & 5 & 512  & 0.2708 & 0.4142 & 0.2117 & 1.5525 & 0.3991 & 0.3695 & 0.7378 & 1.1428 & 0.6588 & 0.117  \\
    \rowcolor{LightGrey}Leverage           & 5 & 2048 & 0.2799 & 0.4207 & 0.2140 & 1.5224 & 0.4033 &  0.3630 & 0.7324 & 1.1204 &  0.6488 & 0.095  \\
    \bottomrule
  \end{tabular}%
  }\vspace{-2ex}
\end{table*}

\vspace{-1ex}
\subsection{Ablation and Sensitivity Study}\vspace{-1ex}
\noindent\textbf{Impact of Checkpoint Selection Strategy.} In Table \ref{tab:ablation_nuscenes_c}, we compare four strategies for choosing $K=5$ checkpoints under heavy Motion Blur: Random sampling, Recent (the latest five), KMeans++ clustering in feature space, and our Leverage‐score ranking.  Random yields a reduced detection mAP of 0.2740, weaker tracking (AMOTA = 0.2152) and planning (CR-Avg = 0.112). Recent performs worst across all tasks (mAP 0.2480, AMOTA 0.1866, CR-Avg 0.149), indicating catastrophic forgetting when only the newest checkpoints are merged. KMeans++ yields a marginal 0.17\% lift in NDS over Random and reduces collision risk by 6.3\%, reflecting its ability to capture diverse feature modes. However, KMeans++ is still outperformed by the proposed method (-3.8\% mAP for detection), highlighting that pure feature clustering cannot match the important informativeness captured by leverage‐score ranking. Overall, the proposed Leverage‐score selection consistently achieves the best results by explicitly identifying the most informative, complementary checkpoints carrying long-term knowledge.

\noindent{\textbf{Impact on Number of Merged Checkpoints.}}  Table~\ref{tab:ablation_nuscenes_c} compares selecting $K\!=\!3$, $5$, or $9$ checkpoints (with $d'\!=\!1024$) for model merging under Motion Blur corruption. With only $K\!=\!3$, detection mAP drops from 0.2851 to 0.2655, and tracking AMOTA falls from 0.2241 to 0.2077, indicating insufficient coverage of knowledge diversity. Increasing to $K\!=\!9$ recovers much of this gap (mAP 0.2818, AMOTA 0.2195) but yields only marginal gains, in mapping mAP (0.4167 vs.\ 0.4103 at $K\!=\!5$). The near‐parity between $K\!=\!5$ and $9$ suggests redundant information beyond five checkpoints. Balancing performance and memory fingerprint, we thus adopt $K\!=\!5$ in all experiments.

\noindent{\textbf{Impact on Dimension of Random Projection.}}  We additionally examine the effect of varying the random projection dimension $d'$ among \{256, 512, 1024, 2048\}. As Table~\ref{tab:ablation_nuscenes_c} shows, at $d=256$, the performance is only slightly below that of $d'=1024$ (mAP 0.2749 vs.\ 0.2851; NDS 0.4176 vs.\ 0.4264), demonstrating that very compact fingerprints still capture most of the critical variability. At $d'=2048$, results nearly match the $d=1024$ but at twice the memory cost. Therefore, $d'=1024$ offers the best trade-off between performance and fingerprint.
\begin{figure*}[t]\vspace{-3ex}
    \includegraphics[width=1\linewidth]{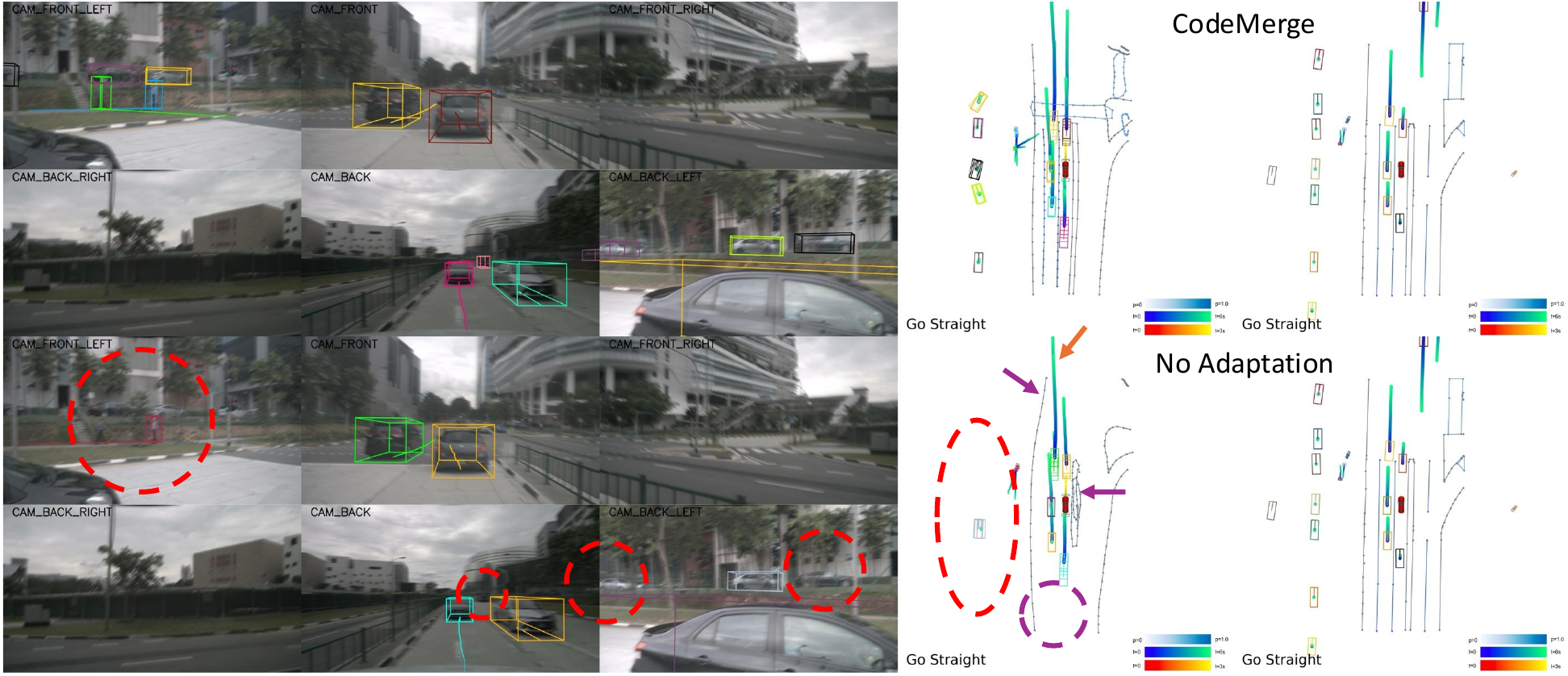}\vspace{-1ex}
    \caption{\textbf{Visualization of outputs of SparseDrive (bottom) and after CodeMerge adaptation (upper) under severe motion blur.} TTA greatly improves detection by capturing more true positive instances, which consequently enhances downstream mapping and planning accuracy (right).\vspace{-3ex} }
    \label{fig:vis}
\end{figure*}

\noindent \textbf{Efficiency Analysis.} We further compare the GPU memory and total TTA runtime of {CodeMerge} \begin{wraptable}{r}{0.4\textwidth}\vspace{-2ex}
  \centering
  \caption{GPU memory (Mb) and total TTA runtime (s) for a single TTA run.\vspace{-1ex}}
  \label{tab:efficiency}
  \resizebox{\linewidth}{!}{%
    \begin{tabular}{l|l c c}
      \toprule
      Model      & Method & GPU Memory & Runtime \\ 
      \midrule
      \multirow{2}{*}{SECOND}      & MOS   & 17,411 & 1,813  \\ 
                                    &\cellcolor{LightGrey} Ours  & \cellcolor{LightGrey}16,041 & \cellcolor{LightGrey}1,054  \\ 
      \midrule
      \multirow{2}{*}{SparseDrive} & MOS   & 39420 & 37,619 \\ 
                                    & \cellcolor{LightGrey}Ours  & \cellcolor{LightGrey}29868 & \cellcolor{LightGrey}27,359 \\ 
      \bottomrule
    \end{tabular}
    }
    \vspace{-2ex}
\end{wraptable} against MOS on both the SECOND and SparseDrive detectors. As reported in Table~\ref{tab:efficiency}, with SECOND, MOS consumes 17.4 GiB and requires 1,813 seconds per adaptation run, whereas CodeMerge uses only 16.0 GiB (\textbf{–8.0\%}) and completes in 1,054 s (\textbf{–41.8\%}). The savings are even more pronounced on SparseDrive: MOS demands 42.7 GiB and 37,619 seconds, while CodeMerge needs just 31.2 GiB (\textbf{–27.0\%}) and 27,359 seconds (\textbf{–27.3\%}). These gains arise from our fingerprint-guided merging, which projects each checkpoint into a compact embedding and computes leverage weights on the fly (requiring only one extra forward pass), rather than loading and forwarding $K$ full models as MOS does. This design drastically reduces memory footprint and latency, making CodeMerge well-suited for real-time autonomous driving applications.

\noindent \textbf{Quantitative Analysis.} We visualize predictions with CodeMerge (top row) against the non-adapted SparseDrive baseline (bottom row) in Figure \ref{fig:vis} to illustrate how on-the-fly merging enhances every stage of the end-to-end pipeline. In detection and tracking, CodeMerge produces tight, correctly aligned 3D boxes, but the baseline suffers a large number of missed or misplaced detections (highlighted in red dashed circles). In mapping, our method reconstructs dense, straight-lane boundaries and curb lines, validating its ability to preserve semantic consistency. In contrast, the baseline yields sparse, crooked lanes and missing curbs (highlighted in purple circles/arrows), degrading map fidelity. Finally, CodeMerge’s planned trajectory remains centered in the lane and safely avoids dynamic objects, while the baseline’s path drifts toward the curb (highlighted in an orange arrow) and even intersects an oncoming track, demonstrating unsafe behavior. In summary, these qualitative results confirm that leveraging compact fingerprints and leverage‐score–guided merging yields better detections, more robust tracking, and safer trajectories under severe real-world corruptions.

\vspace{-2ex}
\section{Conclusion}\vspace{-2ex}
% In this work, we address the challenge of domain shifts in lidar-based and vision-centric detection under extreme scenarios by introducing CodeMerge, a highly robust and efficient model-merging framework. CodeMerge demonstrates superior performance in the face of various sensor malfunctions and adverse weather conditions, highlighting its critical importance for ensuring autonomous driving safety under extreme conditions. CodeMerge operates by constructing a compact key-value codebook within a latent parameter vector space, enabling efficient model merging while effectively preventing interference among model parameters. By leveraging a robust multi-temporal-spatial model ensemble for self-supervision of pretrained models, CodeMerge adapts dynamically during test time to diverse extreme environments, achieving exceptional robustness. Future work will integrate large language models and vision-language models to further enhance the safety and reliability of vision-centric end-to-end autonomous driving algorithms under extreme environmental conditions.

In this work, we address the challenge of online adaptation to domain shifts for both LiDAR-based and vision-centric end-to-end AD detection under extreme conditions. Our proposed CodeMerge framework effectively mitigates cross-dataset and corruption-induced distribution shifts, while reducing GPU memory consumption and inference latency by approximately 27\% compared to state-of-the-art TTA methods. Notably, other downstream modules such as mapping and planning receive performance improvements without task-specific fine-tuning due to enhanced detection outputs. However, this study represents an early attempt to address robustness in end-to-end AD, and major experiments have been primarily conducted on the SparseDrive architecture. The primary bottleneck remains that popular architectures, such as UniAD and VAD, experience over tenfold performance degradation on nuScenes-C, hindering effective adaptation training. Future work will investigate strategies to further accelerate adaptation and enhance robustness under dynamic driving conditions.
% ---- Bibliography ----
\bibliographystyle{plain}
\bibliography{main_bib.bib}

%%%%%%%%%%%%%%%%%%%%%%%%%%%%%%%%%%%%%%%%%%%%%%%%%%%%%%%%%%%%

\appendix

\section{Technical Appendices and Supplementary Material}
We include additional technical details in the following appendices:
\begin{itemize}
    \item Section~\ref{sec:implementation} (Implementation Details): Describes the full experimental setup, including training schedules and hyperparameter configurations.
    \item Section~\ref{sec:metrics} (Evaluation Metrics): Provides definitions and explanations for all evaluation metrics used across detection, tracking, mapping, and planning tasks.
    \item Section~\ref{sec:supp_vis} (Additional Visualizations): Presents qualitative results and visual comparisons illustrating the adaptation performance of the end-to-end AD system under various distribution shifts.
    \item Section~\ref{sec:related} (Related Work): Summarizes the relevant literature in TTA and model merging.
\end{itemize}

\subsection{Implementation Details}\label{sec:implementation}
% Our framework is built upon \textsc{SparseDrive-S}~\cite{SparseDrive} for evaluating end-to-end autonomous driving tasks and \textsc{OpenPCDet}~\cite{openpcdet2020} toolkit for LiDAR-based 3D object detection. All experiments are conducted using a single NVIDIA RTX A6000 GPU with 48\,GB of memory. For the end-to-end task, ResNet50 is used as backbone module, 
% We use a batch size of \texttt{1} for autonomous driving and \texttt{8} for 3D detection, and apply consistent hyperparameter settings of $A = \texttt{XX}$ and $B = \texttt{XX}$ across all tasks. The learning rate

For the end-to-end autonomous driving task, we employ ResNet50 \cite{ResNet} as the backbone network to uniformly process image data from both nuScenes and nuScenes-C \citep{DBLP:journals/pami/XieKZRPCL25} datasets. All input images are resized to 256×704. We use a 900×256 instance query as input to the transformer layers. Our optimization strategy utilizes the AdamW optimizer, configured with a weight decay of 0.001 and an initial learning rate of $1\times10^{-7}$. To balance computational efficiency and prediction accuracy, we apply a random projection module to reduce the dimensionality of query features extracted from the pretrained model, resulting in a compact 1024-dimensional feature vector, and manage predictions through a model bank with a limited capacity of five models. Through self-supervised training on detection and tracking heads, the model accurately predicts ten classes as well as the associated instance IDs. For the point cloud detection tasks, we adopt the SECOND \cite{DBLP:journals/sensors/YanML18} as our pretrained model. We configure the training with a batch size of 8, a learning rate of 0.01, and a weight decay of 0.01. Additionally, we utilize a 900×256 dimensional 3D feature vector as input to the leverage module, enabling efficient and effective model merging.

\subsection{Evaluation Metrics in End-to-End AD}\label{sec:metrics}
We follow standard evaluation protocols to assess each task module for end-to-end AD system.
\noindent\textbf{Detection Metrics.}
We use nuScenes metrics including mean Average Precision (mAP) and five error-based scores: mean Average Translation Error (mATE), Scale Error (mASE), Orientation Error (mAOE), Velocity Error (mAVE), and Attribute Error (mAAE). Together, they evaluate spatial, geometric, and semantic aspects of 3D box predictions. The nuScenes Detection Score (NDS) aggregates these metrics into a single score for holistic performance evaluation.

\noindent\textbf{Tracking Metrics.}
Tracking performance is measured using Average Multi-Object Tracking Accuracy (AMOTA), Precision (AMOTP), and Recall. These metrics capture association quality, localization precision, and coverage of tracked instances.

\noindent\textbf{Online Mapping Metrics.}
We compute class-wise Average Precision (AP) for static map elements (e.g., lane dividers, crossings, road boundaries) and report mean AP across categories to reflect mapping accuracy and consistency.

\noindent\textbf{Motion Prediction Metrics.}
We evaluate prediction with best-of-$K$ trajectory metrics: minimum Average Displacement Error (minADE), minimum Final Displacement Error (minFDE), and Miss Rate (MR). We also report End-to-end Prediction Accuracy (EPA), which reflects cascading errors across detection, tracking, and forecasting stages.
\noindent\textbf{Planning Metrics.}
We assess planning quality using two key indicators: collision rate, which measures the frequency of collisions during trajectory execution, and L2 distance to goal, which quantifies the Euclidean distance between the final position and the intended goal. Together, these metrics reflect the safety and goal-reaching accuracy of the planned motion.

\subsection{More Visualizations}\label{sec:supp_vis}

\begin{figure}[htbp]
  \centering
  % 第一张
  \includegraphics[width=0.99\linewidth]{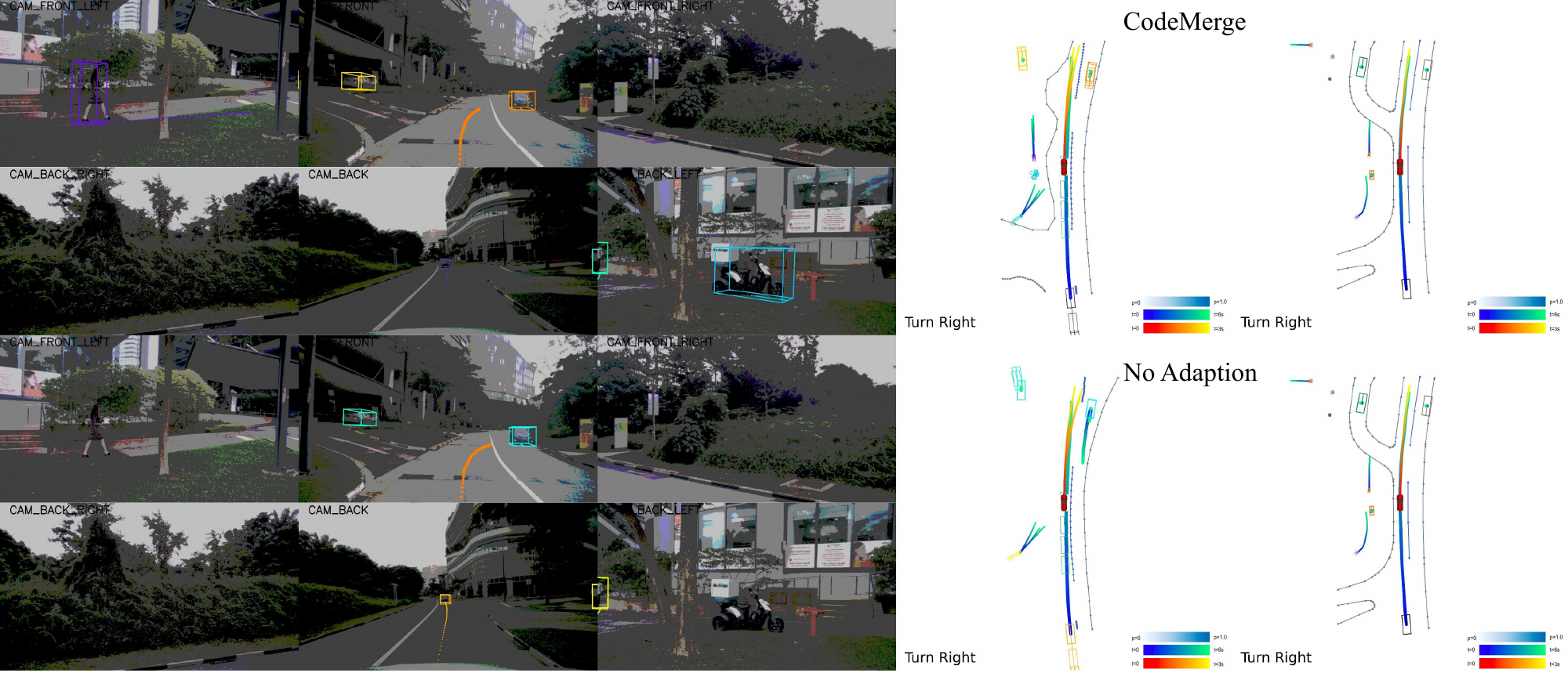}\\[1ex]
  % 第二张
  \includegraphics[width=0.99\linewidth]{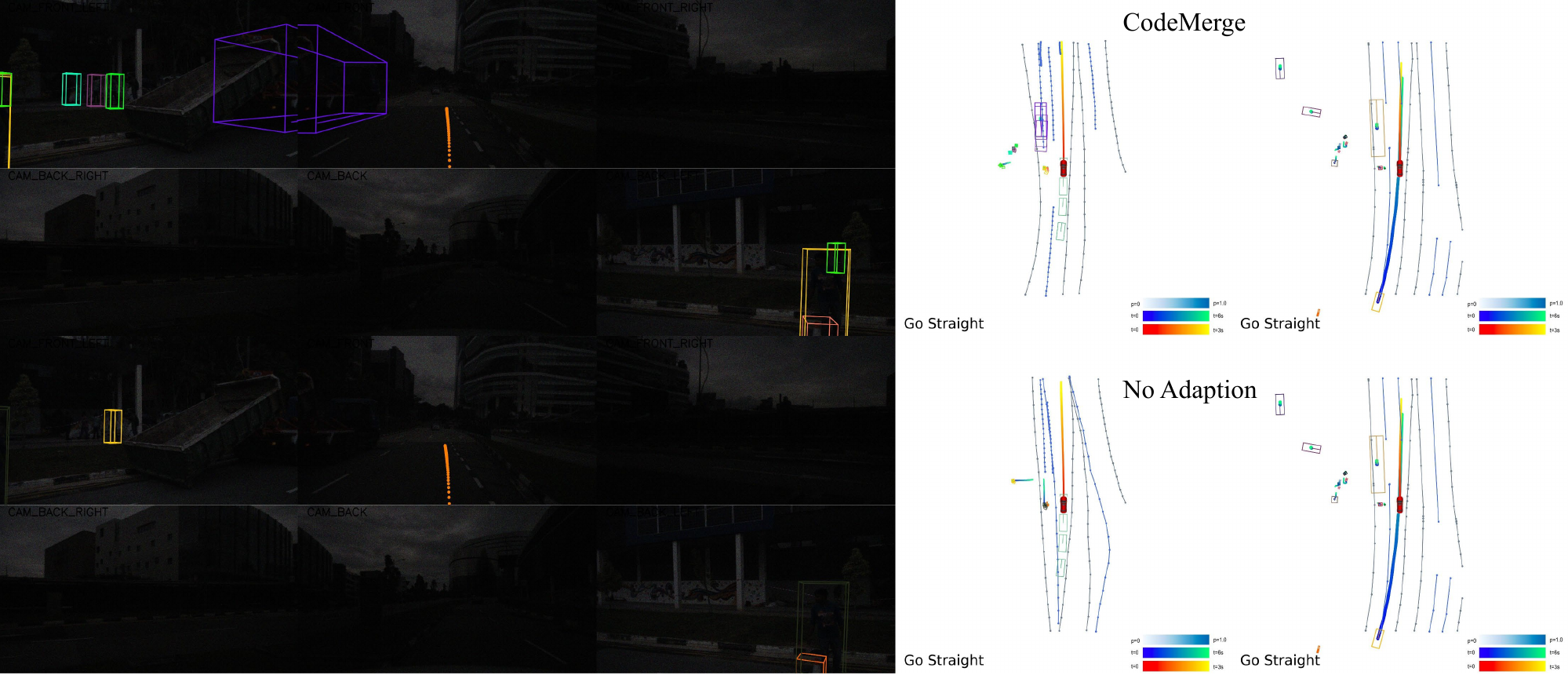}\\[1ex]
  % 第三张
  \includegraphics[width=0.99\linewidth]{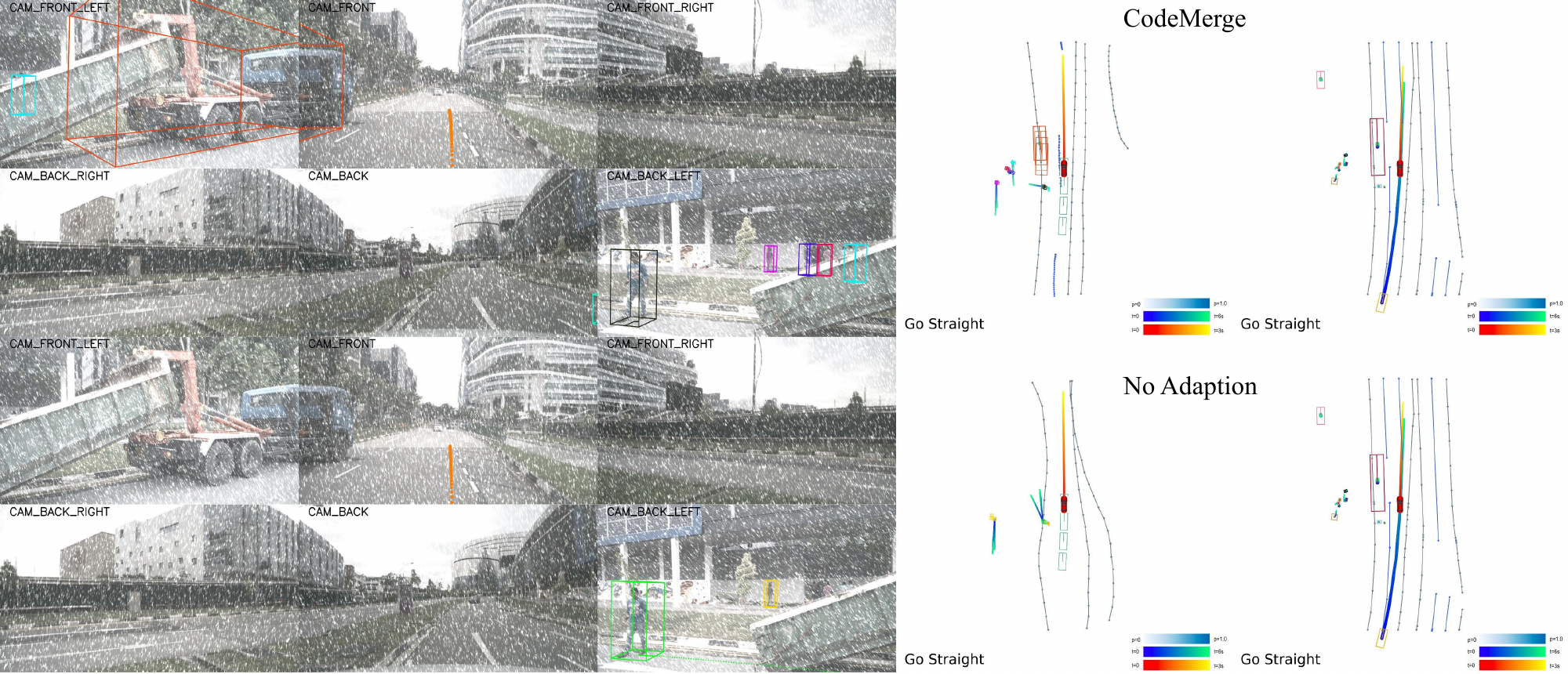}
  \caption{Visualization of outputs of SparseDrive (bottom) and after CodeMerge adaptation (upper) under severe ColorQuant, LowLight, and Snow. TTA greatly improves detection by capturing more true positive instances.}
  \label{fig:more_vis}
\end{figure}

In Figure~\ref{fig:more_vis}, we present additional visualized predictions from both the non-adapted SparseDrive and the SparseDrive model adapted at test-time using the proposed CodeMerge, illustrating performance across a broader range of corruptions.

%%%%%%%%%%%%%%%%%%%%%%%%%%%%%%%%%%%%%%%%%%%%%%%%%%%%%%%%%%%%

\subsection{Related Work}\label{sec:related}
\textbf{Test-time adaptation (TTA)} dynamically updates models during deployment to mitigate distribution shifts~\cite{DBLP:journals/ijcv/LiangHT25, DBLP:journals/ijcv/WangLZCWH25}. Early approaches primarily focus on tuning BatchNorm layers via entropy minimization, adaptive moment estimation, global statistic alignment, or loss landscape smoothing~\cite{DBLP:conf/iclr/WangSLOD21,DBLP:conf/cvpr/MirzaMPB22a,DBLP:conf/cvpr/YuanX023,zhao2023delta,DBLP:conf/iclr/Niu00WCZT23,gong2024sotta,DBLP:conf/icml/NiuMCWZ24}. Subsequent methods explore self-training with confidence-filtered pseudo-labels~\cite{goyal2022test,10107906,zeng2024rethinking}, feature-level consistency or contrastive regularization~\cite{chen2022contrastive,jung2023cafa,DBLP:conf/cvpr/WangZYZL23,DBLP:conf/eccv/ShimKY24,DBLP:conf/eccv/ZouQLKHCJ24,DBLP:conf/cvpr/WangCHHRRALPH24}, robustness through data augmentation~\cite{DBLP:conf/nips/ZhangLF22,DBLP:conf/cvpr/TsaiCCYSSK24}, and leveraging guidance from language models~\cite{DBLP:conf/cvpr/KarmanovGLEX24}. Extending TTA to more challenging perception tasks (\textit{e.g.}, image- or LiDAR-based object detection \cite{DBLP:conf/iccv/LuoCF0BH23, DBLP:conf/iclr/LuoCWYHB23, DBLP:journals/corr/abs-2310-10391}), {MemCLR} aligns 2D detector features using a memory-augmented teacher–student framework~\cite{DBLP:conf/wacv/VSOP23}; {DPO} stabilizes LiDAR-based detection via dual perturbation optimization \citep{DBLP:conf/mm/ChenWL0H24}; {Reg-TTA3D} generates noise-consistent pseudo-labels to supervise low-confidence 3D boxes using high-confidence ones~\cite{DBLP:conf/eccv/YuanZGYSQC24}; and {MOS} enhances adaptation stability by dynamically merging the top-$K$ diverse checkpoints for supervision~\cite{DBLP:conf/iclr/ChenMB0HL25}. Despite their effectiveness, existing methods adapt only perception tasks, while adapting unified end-to-end autonomous driving systems at test time remains unexplored.

\noindent\textbf{Model Merging} studies how weight-space operations can effectively compose, refine, or repair vision models through checkpoint averaging, gradient matching, or arithmetic edits to task-specific weight vectors~\cite{DBLP:conf/emnlp/MorrisonSHKDD24, DBLP:conf/icml/WortsmanIGRLMNF22, DBLP:conf/iclr/DaheimMPGK24, DBLP:conf/iclr/IlharcoRWSHF23, DBLP:conf/iclr/AinsworthHS23,DBLP:conf/nips/YadavTCRB23}. Recent literature highlights the effectiveness of these merging techniques in enhancing generalization across tasks such as zero-shot learning~\cite{DBLP:conf/cvpr/WortsmanIKLKRLH22}, open-set learning~\cite{DBLP:conf/nips/QuHCL23}, domain adaptation and generalization~\cite{DBLP:conf/nips/RameKRRGC22, DBLP:conf/nips/ArpitWZX22, DBLP:conf/iclr/DinuHBNHEMPHZ23}, and cross-domain tasks involving 3D LiDAR point clouds~\cite{DBLP:conf/nips/Jiang0LZWMCZL24, DBLP:conf/iclr/ChenMB0HL25}. In this work, we build upon the strengths of model merging techniques to enable efficient, on-the-fly adaptation within end-to-end autonomous driving pipelines.

\newpage

\end{document}